%% file: main.tex
\documentclass[10pt,twocolumn,letterpaper]{article}

\usepackage[pagenumbers]{cvpr} 

\usepackage{multirow}

\usepackage{algorithm}
\usepackage{algpseudocode}

\algtext*{EndIf} 
\algtext*{EndFor} 
\algtext*{EndWhile} 

\usepackage{makecell}
\usepackage{arydshln}
\usepackage[most]{tcolorbox} 
\usepackage{array}
\usepackage{tabularx} 
\usepackage{caption} 

\usepackage{url}
\usepackage{graphicx}
\usepackage{xspace}
\usepackage{multirow,marvosym}

\usepackage{xcolor}

\newcommand\blfootnote[1]{%
  \begingroup
  \renewcommand\thefootnote{}\footnote{#1}%
  \addtocounter{footnote}{-1}%
  \endgroup
}
\usepackage{amsmath}


\usepackage{lineno}     

\definecolor{cvprblue}{rgb}{0.21,0.49,0.74}
\usepackage[pagebackref,breaklinks,colorlinks,allcolors=cvprblue]{hyperref}

\def\paperID{4047} 
\def\confName{CVPR}
\def\confYear{2025}

\title{VLM-R1: A Stable and Generalizable R1-style Large Vision-Language Model}

\author{%
    Haozhan Shen$^{1}$, 
    Peng Liu$^{2}$, 
    Jingcheng Li$^{2}$, 
    Chunxin Fang$^{2}$, 
    Yibo Ma$^{2}$, 
    Jiajia Liao$^{2}$,  
    Qiaoli Shen$^{2}$, \\
    Zilun Zhang$^{1}$,
    Kangjia Zhao$^{1}$, 
    Qianqian Zhang$^{2}$,
    Ruochen Xu$^{2}$,
    Tiancheng Zhao$^{2,3}$\textsuperscript{\Letter} \\[2mm]
    $^1$ Zhejiang University \quad
    $^2$ Om AI Research \quad
    $^3$ Binjiang Institute of Zhejiang University \\
    {\tt\small \{tianchez\}@zju-bj.com} \\
}
\begin{document}

\twocolumn[{
\maketitle
\vspace{-20pt}
\begin{center}
    \includegraphics[width=1\linewidth]{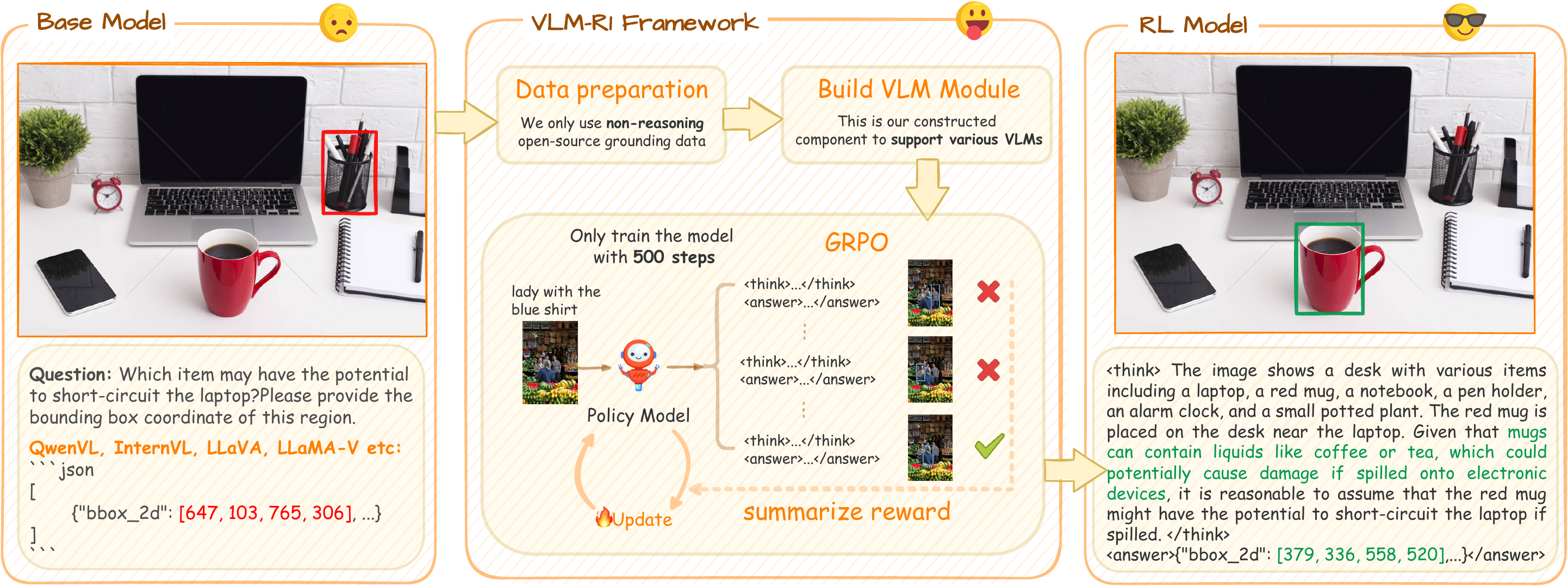}
    \captionof{figure}{VLM-R1 provides a standard pipeline to enhance base vision-language models (VLMs) with reinforcement learning.\\\vspace{2\baselineskip} 
    }
    \label{fig:case}
\end{center}
}]

\blfootnote{\textsuperscript{\Letter} Corresponding author.}

\input{sec/0_abstract}
\input{sec/1_intro}

\input{sec/2_related}
\input{sec/3_vlm-r1}
\input{sec/4_reward}
\input{sec/5_experiment}
\input{sec/6_discussion}

\section{Conclusion}
In this work, we introduce VLM-R1, the unified framework that brings R1-style reinforcement learning into the domain of visual understanding. Our framework is tailored to vision-language models and supports flexible data definition, model modularity, and training scalability. Using VLM-R1, we successfully apply RL to two representative visual understanding tasks---referring expression comprehension and open-vocabulary object detection---demonstrating substantial gains in both task performance and out-of-domain generalization.
Beyond empirical results, we provide practical insights into reward engineering, data selection, and model scaling that are critical for applying RL effectively to complex vision-language tasks. Our work lays the foundation for broader adaption of reinforcement learning in vision-language research. In future work, we aim to explore cross-task generalization and extend VLM-R1 to more challenging multimodal scenarios.

\newpage

{
    \small
    \bibliographystyle{ieeenat_fullname}
    \bibliography{main}
}



\end{document}

%% file: sec/0_abstract.tex
\begin{abstract}
Recently, DeepSeek R1 has shown that reinforcement learning (RL) can substantially improve the reasoning capabilities of Large Language Models (LLMs) through a simple yet effective design. The core of R1 lies in its rule-based reward formulation, which leverages tasks with deterministic ground-truth answers to enable precise and stable reward computation. In the visual domain, we similarly observe that a wide range of visual understanding tasks are inherently equipped with well-defined ground-truth annotations. This property makes them naturally compatible with rule-based reward mechanisms. Motivated by this observation, we investigate the extension of R1-style reinforcement learning to Vision-Language Models (VLMs), aiming to enhance their visual reasoning capabilities. To this end, we develop VLM-R1, a dedicated framework designed to harness RL for improving VLMs’ performance on general vision-language tasks. Using this framework, we further explore the feasibility of applying RL to visual domain. Experimental results indicate that the RL-based model not only delivers competitive performance on visual understanding tasks but also surpasses Supervised Fine-Tuning (SFT) in generalization ability. Furthermore, we conduct comprehensive ablation studies that uncover a series of noteworthy insights, including the presence of reward hacking in object detection, the emergence of the \textit{“OD aha moment”}, the impact of training data quality, and the scaling behavior of RL across different model sizes. Through these analyses, we aim to deepen the understanding of how reinforcement learning enhances the capabilities of vision-language models, and we hope our findings and open-source contributions will support continued progress in the vision-language RL community. Our code and model are available at \href{https://github.com/om-ai-lab/VLM-R1}{https://github.com/om-ai-lab/VLM-R1}.
\end{abstract}

%% file: sec/1_intro.tex
\section{Introduction}\label{sec:intro}
The introduction of OpenAI o1~\cite{jaech2024openai} demonstrated that reinforcement learning (RL), which enables Large Language Models (LLMs) to directly learn from feedback on their outputs, can significantly enhance their reasoning capabilities. More recently, DeepSeek R1~\cite{guo2025deepseek} further advanced this insight by showing that simple rule-based rewards, without the need for a separate learned reward model~\cite{liu2024skywork,ouyang2022training,zang2025internlm}, are sufficient to autonomously equip LLMs with complex reasoning performance.

A key factor behind this success is that the rule-based reward design is easily applicable to tasks with deterministic ground-truth answers, allowing for stable and interpretable reward signals. In the visual domain, analogously, there exist numerous visual understanding tasks inherently including precise and objectively defined ground-truth annotations. For example, tasks such as Referring Expression Comprehension~(REC)~\cite{mao2016generation,yu2016modeling} can directly adopt Intersection-over-Union (IoU) between the predicted bounding box and the ground-truth annotation as an explicit reward metric. Motivated by these observations, it becomes intuitive to investigate whether similar RL methodologies can comparably enhance the reasoning capabilities of Vision-Language Models~(VLMs).


To this end, we develop \textbf{VLM-R1}, a dedicated and extensible framework designed to apply RL to improve the performance of VLMs on general vision-language tasks. VLM-R1 is built with flexibility, scalability, and ease of experimentation in mind. It supports a wide range of configurations and is tailored for research on RL-based optimization in the context of VLMs. Key features of VLM-R1 include:
\begin{itemize}
    \item \textit{GRPO Compatibility}: Fully supports the native GRPO~\cite{shao2024deepseekmath} algorithm with fine-grained control over all hyperparameters.
    \item \textit{LoRA-based Training}: Enables parameter-efficient training via LoRA~\cite{hu2022lora}, suitable for limited-resource settings.
    \item \textit{Multi-node Training}: Supports distributed training across multiple GPUs or server nodes for scalability.
    \item \textit{Multi-image Input}: Supports multiple images per sample, facilitating complex multi-image reasoning tasks.
    \item \textit{Model Flexibility}: Compatible with various VLMs, currently supporting QwenVL~\cite{bai2025qwen2,wang2024qwen2} and InternVL~\cite{chen2024internvl,chen2024expanding}.
    \item \textit{Custom Dataset Support}: Easily integrates user-defined datasets, allowing for task-specific or domain-specific experiments
    \item \textit{Mixed Modality Training}: Supports training on both image-text and pure-text datasets, including hybrid combinations.
\end{itemize}

By providing a unified, modular, and highly adaptable training pipeline, VLM-R1 serves as a powerful tool for advancing research at the intersection of reinforcement learning and vision-language modeling.

\begin{table}[t!]
\footnotesize
  \begin{center}
  \resizebox{\hsize}{!}{
  \setlength\tabcolsep{1pt}
    \begin{tabular}{l c c c c c c c c} 
    \toprule
        \textbf{Model} &  & Model Size &  & Refcoco$_{val}$ & Refcoco+$_{val}$ & Refcocog$_{val}$ & & ODinW \\
    \midrule
        Qwen2.5-VL-3B & \vline & 3.75B & \vline & 89.1 & 82.4 & 85.2 &\vline & 37.5\\
        Grounding DINO & \vline & 341M & \vline & \textbf{90.6} &\textbf{88.2} & \textbf{86.1} &\vline & \textbf{55.0} \\
    \bottomrule
    \end{tabular}
    }
  \end{center}
  \caption{Performance comparison between Qwen2.5-VL-3B and Grounding DINO on REC and OVD tasks. Even though having over 10× the number of Grounding DINO, Qwen2.5-VL-3B still falls short on these evaluation datasets. It shows the drawback of VLMs on these visual understanding tasks.}
  \label{table:model compare}
  \vspace{-3mm}
\end{table}

In this report, utilizing VLM-R1, we select two visual understanding tasks --- Referring Expression Compression~(REC) and Open-Vocabulary Object Detection~(OVD) --- to explore the feasibility and effectiveness of applying RL to VLMs. REC and OVD share a common output format—bounding boxes—but differ significantly in task complexity. In REC, the model is expected to predict a single bounding box based on a given query, whereas in OVD, the model must accurately output each corresponding bounding box for every queried target. This contrast allows us to analyze how tasks with similar output structures but varying difficulty levels influence the effectiveness of reinforcement learning in VLMs. Moreover, we observe that VLMs often underperform compared to specialized vision models~(e.g., Grounding DINO~\cite{liu2024grounding,ren2024grounding}, OmDet~\cite{zhao2022omdet,zhao2024omdet}) on these tasks. As shown in Table~\ref{table:model compare}, despite having over 10$\times$ the number of parameters as Grounding DINO, Qwen2.5-VL-3B still lags behind in performance on both REC and OVD benchmarks. This performance gap raises an important question: can reinforcement learning be leveraged to enhance VLMs’ effectiveness on these challenging visual understanding tasks?

The experimental results demonstrate that RL substantially improves visual understanding performance in VLMs compared to supervised fine-tuning (SFT), and, more importantly, yields significantly greater gains in generalization ability on complicated, real-world benchmarks. In the context of REC, our 3B RL model achieves a score of 63.16 on the out-of-domain evaluation benchmark LISA-Grounding~\cite{lai2024lisa}(vs. 54.82 for SFT). For OVD task, the 3B RL model reaches 21.1 AP on COCO~\cite{lin2014microsoft}(vs. 17.8 for SFT; 14.2 for 7B baseline model), and new SOTA 31.01 nms-AP on OVDEval~\cite{yao2023evaluate}(vs. 26.50 for SFT; 29.08 for 7B model), especially excelling in complex sub-tasks.

In addition, comprehensive ablation studies further uncover a range of important insights. For instance, we observe the reward hacking in object detection, and conduct reward engineering to alleviate it, where the model emerges \textit{``OD aha moment"}, first reasoning about object presence before predicting. Furthermore, we also demonstrate that the careful selection of training data could improve the final performance, and analyze the impact of model size. Taken together, our findings suggest that more complex tasks---such as open-vocabulary object detection---demand additional optimization to achieve strong performance, whereas relatively simpler tasks like REC can be effectively addressed with fewer modifications.
Our contributions could be summarized as:
\begin{itemize}
    \item We develop \textbf{VLM-R1} based on open-r1, a dedicated and extensible framework designed to apply reinforcement learning to improve the performance of vision-language models, aiming for flexibility, scalability, ease of experimentation, and the supporting a wide range of RL configurations.
    \item We demonstrate the effectiveness of applying reinforcement learning to vision-language models through training two essential visual understanding tasks: referring expression compression and open-vocabulary object detection. Trained with VLM-R1, our RL model achieves a performance improvement compared to the SFT counterpart, especially on the complicated, real-world out-of-domain benchmarks.
    \item Our extended ablation studies reveal a series of interesting insights, including the presence of reward hacking in object detection, the emergence of \textit{``OD aha moment"}, the influence of training data quality, and the RL effects across model scales. We report these insights and analyze how to well-tune reinforcement learning to enhances the performance of VLMs.
    \item We release the framework codebase and all model weights, with the hope of contributing to the open-source community in vision-language reinforcement learning.
\end{itemize}


%% file: sec/2_related.tex
\section{Related Work}

\subsection{Vision-Language Models}

Since the advent of large language models\;(LLMs), they have achieved success in various linguistic applications, facilitating the emergence of Vision-Language Models~(VLMs), with pioneering works including \cite{alayrac2022flamingo,li2023blip,koh2023grounding}. Following these, LLaVA~\cite{liu2024visual} employed GPT-4~\cite{achiam2023gpt} to develop training data and achieve promising performance in visual dialogue and visual reasoning, inspiring a series of works focused on visual instruction data~\cite{liu2024improved,instructblip,chen2023sharegpt4v}. However, a key limitation of the VLMs at that time lies in their constrained image input resolution, which is restricted by the capabilities of their underlying vision encoders~\cite{radford2021learning,EVA-CLIP,zhai2023sigmoid}. To overcome this, the AnyRes mechanism was introduced~\cite{liu2024llavanext,chen2024internvl,chen2024dragonfly}, allowing flexible handling of images with varying resolutions and aspect ratios. This advancement improves the perceptual capacity of VLMs for diverse visual inputs and further enhances their reasoning abilities. Today, some of the most widely adopted open-source VLM series include LLaVA\cite{liu2024visual,liu2024llavanext,li2024llava}, QwenVL\cite{wang2024qwen2,bai2025qwen2}, and InternVL~\cite{chen2024expanding,chen2024internvl}.


\subsection{Attempts of applying R1 to VLMs}
Several concurrent studies have explored the application of R1 to Vision-Language Models (VLMs). Concurrent work R1-OneVision~\cite{yang2025r1} and R1-V~\cite{chen2025r1v} are among the notable works in this direction. R1-OneVision proposed a cross-modal reasoning pipeline that converts images into visual formal representations, which are then used to construct a visual reasoning dataset via a language model. The VLM is first trained on this dataset, followed by an RL stage to further enhance its reasoning capability. In parallel, R1-V introduced the GRPO method~\cite{shao2024deepseekmath} from DeepSeek R1~\cite{guo2025deepseek} into VLM training, targeting object-counting tasks, and remarkably enabled a 3B model to outperform a 72B model. Soon afterward, VisualThinker-R1-Zero~\cite{zhou2025r1} was presented, which demonstrated that applying R1 to base VLM instead of an instruction-tuned model could achieve more considerable performance improvements and successfully trigger the emergence of the so-called ``visual aha moment". Another work observing the appearance of the aha moment and the increasing length of the model response that is akin to the phenomena in DeepSeek R1 is MM-Eureka~\cite{meng2025mm}, which applied RLOO~\cite{kool2019buy,ahmadian2024back} to both the 8B instruction-tuned VLM and the 38B base VLM. Analogous to R1-OneVision, Vision-R1~\cite{huang2025vision} constructed a multimodal CoT dataset in terms of converting vision information to a language format and feeding it into a language reasoning model. This dataset serves as the cold start training data followed by the GRPO to further strengthen the multimodal reasoning ability of the model. In addition, Curr-ReFT\cite{deng2025boosting} proposed a three-stage reinforcement learning with progressive difficulty-level reward to optimize RL training, and LMM-R1\cite{peng2025lmm} presented a two-stage rule-based RL, where they first adopted text-only data to strengthen the basic reasoning abilities of the model and then continued RL on limited complex multimodal reasoning tasks.

Most of the above studies focus mainly on improving performance in multimodal mathematics tasks~\cite{lu2023mathvista,zhang2024mathverse,wang2024measuring}. In contrast, Visual-RFT~\cite{liu2025visual} applies RL to visual perception tasks, making it more closely related to our work. However, our study provides a more comprehensive investigation, going beyond a simple comparison between supervised fine-tuning (SFT) and RL. Specifically, we further analyze the role of reward engineering and systematically examine the impact of careful training data selection, particularly for complex tasks.

%% file: sec/3_vlm-r1.tex
\section{VLM-R1 Framework}

\begin{figure*}[ht]
    \centering
    \includegraphics[width=17cm]{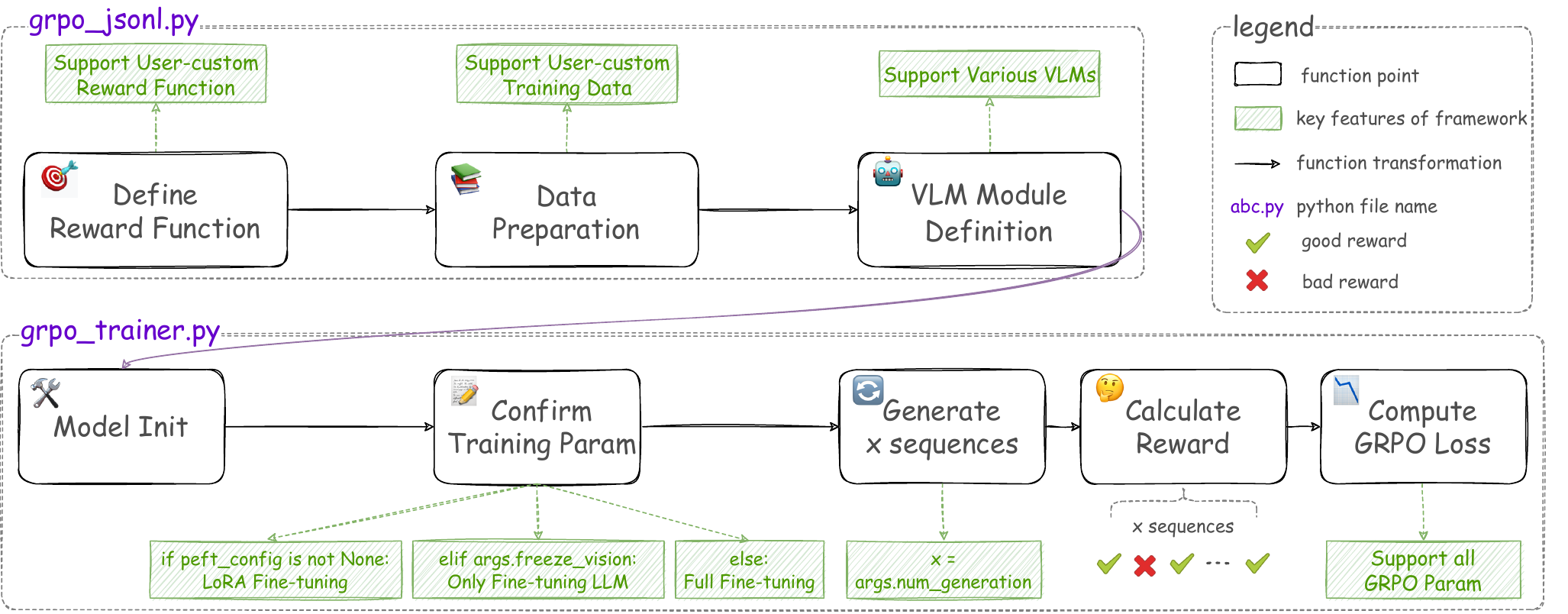}
     \caption{Flow chart of VLM-R1 framework. This chart exhibits the function transformation of the framework. The key features of VLM-R1 are displyed by the \textcolor[RGB]{34,139,34}{green} rectangle.}
    \label{fig:flow chart}
\end{figure*}

In this section, we provide a brief introduction to the proposed VLM-R1 framework. VLM-R1 is built upon Open-R1~\cite{openr1}, an open-source framework for reproducing the language reasoning capabilities of DeepSeek R1. We extend its implementation to the vision-language domain. In addition to ours, there are several other open source frameworks that target vision language reinforcement learning~\cite{zheng2025easyr1,chen2025r1v}. It should be noted that our primary objective is to adapt the R1-style methodology for Vision-Language Models (VLMs). Therefore, our current implementation focuses exclusively on the GRPO~\cite{shao2024deepseekmath} algorithm, as originally adopted by DeepSeek R1. Consequently, VLM-R1 currently supports only GRPO, with plans to integrate additional RL algorithms in future work. In the following, we first present an overview of the framework, followed by a detailed description of the VLM Module, which enables seamless support for various VLM architectures.

\subsection{Overview}\label{sec: framework overview}
As shown in Figure~\ref{fig:flow chart}, the VLM-R1 framework is composed of two main components: grpo\_jsonl.py and grpo\_trainer.py, which together form a complete pipeline for GRPO~\cite{shao2024deepseekmath} algorithm to VLMs. 

In the first stage (grpo\_jsonl.py) serving as a preparation stage, users can flexibly define custom reward functions and prepare training data tailored to their tasks. The framework also supports various VLMs through a modular VLM Module Definition, which will be described in §~\ref{sec:vlm module}. The second stage (grpo\_trainer.py) manages the GRPO training process. It begins with model initialization, followed by confirmation of training parameters decided by the user-custom parameters. We support LoRA fine-tuning, vision tower freezing training, and full parameters training, which could be adapted to distinct compute resources and task requirements. The model subsequently generates multiple sequences, which are scored using the defined reward function. These reward signals are then used to compute the GRPO loss for parameter optimization. 

VLM-R1 provides full support for GRPO training while offering flexibility in reward design, model selection, and optimization strategies, making it a versatile tool for RL-based vision-language research.  
 
\begin{figure}[t!]
    \centering
    \includegraphics[width=8cm]{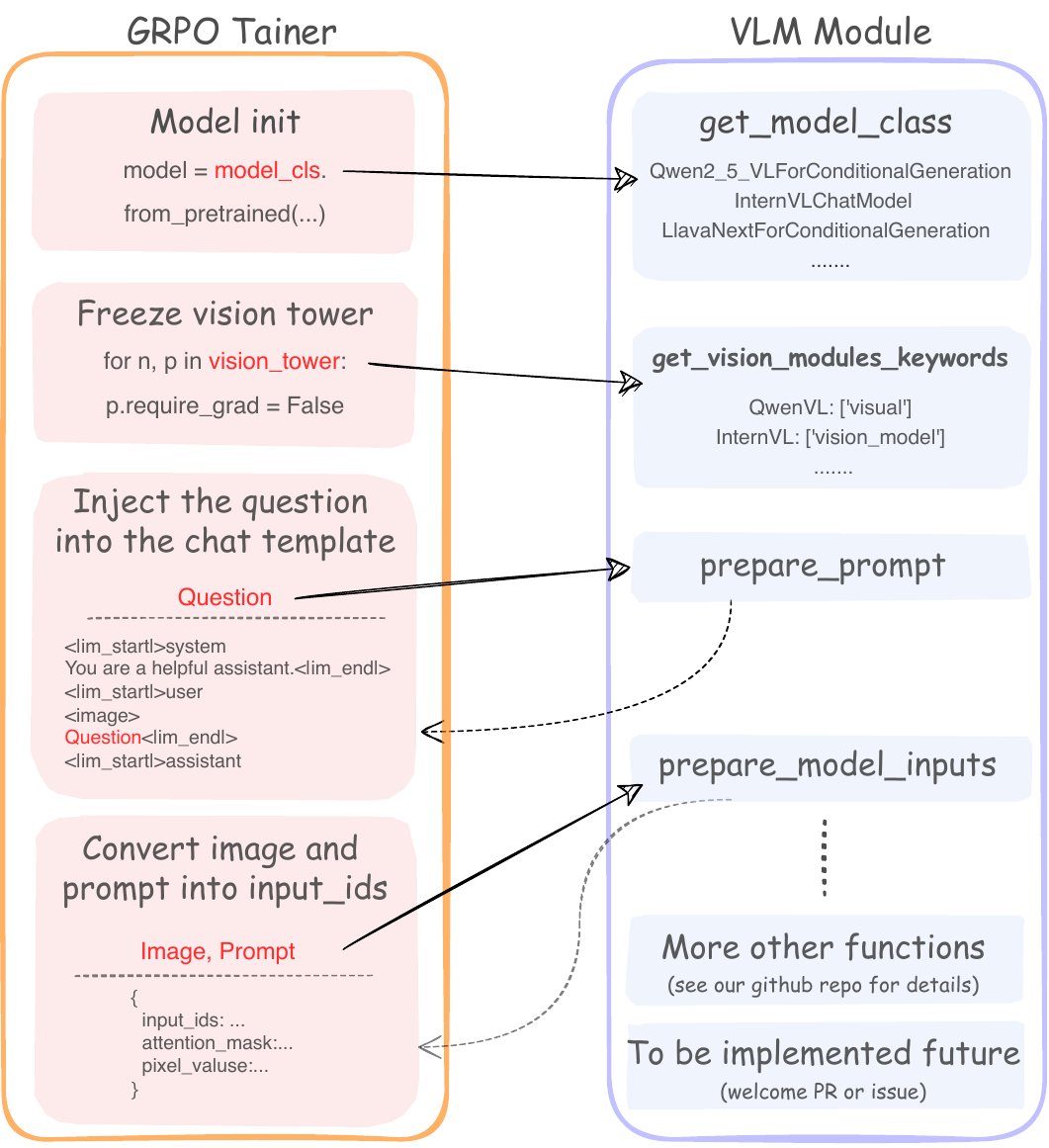}
    \caption{The interaction between Trainer and VLM Module. With the VLM Module, the GRPOTrainer can interact with different VLMs by simply invoking the standardized interfaces without the need to handle model-specific implementations.}
    \label{fig:vlm module}
    \vspace{-2mm}
\end{figure}

\subsection{VLM Module}\label{sec:vlm module}
To facilitate the seamless integration of various VLMs into the training process, we design a unified component, which we refer to as VLM Module. This module encapsulates general VLM functionalities, such as retrieving the model’s class name and formatting input questions into the model-specific chat template. By abstracting these operations, the GRPOTrainer can interact with different VLMs by simply invoking the standardized interfaces provided by the VLM Module, without the need to handle model-specific implementations. This design not only streamlines the integration of new models but also enhances the modularity and readability of the overall framework. The interaction between Trainer and the VLM Module is shown in Figure~\ref{fig:vlm module}.

%% file: sec/4_reward.tex
\section{Reward Design}
As discussed in §~\ref{sec:intro}, we select referring expression comprehension~(REC) and open-vocabulary object detection~(OVD) as representative tasks due to two considerations. 
First, both tasks share a common bounding box output format but differ in complexity, providing a suitable setting to examine the impact of RL across tasks with varying difficulty. 
Second, specialized vision models have consistently outperformed VLMs on these benchmarks, offering a valuable opportunity to assess whether RL can help close this performance gap.

In this section, we first brief the general GRPO algorithm and then introduce the reward design for REC and OVD tasks that be integrated into the GRPO.

\subsection{Abstraction of GRPO}
Unlike reinforcement learning algorithms such as PPO~\cite{schulman2017proximal}, which require an additional critic model to estimate policy performance, Group Relative Policy Optimization (GRPO) directly compares groups of candidate responses, eliminating the need for a separate critic. 
Given a question $q$, GRPO samples $N$ candidate responses $\{o_1, o_2, \ldots, o_N\}$ from the policy $\pi_\theta$ and evaluates each response $o_i$ using a reward function $R(q, o_i)$, which measures the quality of the candidate in the context of the given question. 
To determine the relative quality of these responses, GRPO normalizes the rewards by computing their mean and standard deviation and subsequently derives the advantage as:

\begin{equation}
    A_i = \frac{r_i - \text{mean}\{r_1, r_2, \ldots, r_N\}}{\text{std}\{r_1, r_2, \ldots, r_N\}}
\end{equation}

where $A_i$ represents the advantage of the candidate response $o_i$ relative to other sampled responses. GRPO encourages the model to generate responses with higher advantages within the group by updating the policy $\pi_\theta$ using the following objective:
\begin{align}
    \mathcal{J}_{GRPO}&(\theta) = \mathbb{E}[{\{o_i\}_{i=1}^N\sim\pi_{\theta_{old}}(q)}] \\
    &\frac{1}{N}\sum_{i=1}^N\left\{\min[s_1\cdot A_i,\ s_2 \cdot A_i]-\beta\mathbb{D}_{KL}[\pi_\theta||\pi_{ref}]\right\} \\
    s_1 & = \quad\quad\frac{\pi_\theta(o_i|q)}{\pi_{\theta_{old}}(o_i|q)} \\
    s_2 & = \text{clip}\left(\frac{\pi_\theta(o_i|q)}{\pi_{\theta_{old}}(o_i|q)},1+\epsilon,1-\epsilon\right)
\end{align}
As mentioned in \ref{sec: framework overview}, all hyperparameters in the above equations are included in our proposed VLM-R1 framework.

Subsequently, we will introduce the reward function $R$ adopted for REC and OVD tasks. Following DeepSeek-R1, we use two types of rewards: accuracy reward and format reward.

\subsection{Reward function for referring expression comprehension}
\noindent\textbf{Accuracy reward}.
Referring expression comprehension (REC) is a task that requires the model to identify the region bounding box of the object described by a referring expression.
Denote $q$ as the input question, $b^*$ as the ground truth bounding box, $o$ as the VLM output sentence, and $f_{rec}$ as the function that is used to extract the bounding box from the output sentence.
The accuracy reward for REC is defined as:
\begin{equation}
    R^{rec}_{acc}(q, o) = \text{IoU}(b^*, f_{rec}(o))
\end{equation}
where \text{IoU} is the intersection over union metric. 
This reward function is designed to encourage the model to generate bounding boxes that closely match the ground truth.

\noindent\textbf{Format reward}. Format reward of REC checks whether the response follows the specified format that require the model has to output the json-style response in the \texttt{<answer>} tag and include a bounding-box~\texttt{(<think>...</think><answer>\{...[x1, y1, x2, y2]...\}</answer>)}, returning 1 or 0 based on compliance.

\subsection{Reward function for open-vocabulary object detection}
\noindent\textbf{Accuracy reward}.
Open-vocabulary object detection (OVD) requires the model to detect the given object labels in the image and output the corresponding bounding boxes and class labels.
This task has a similar output format as REC, but is more complex due to the need for both bounding box and class label generation. 
In this task, we prompt the VLM to output a list of bounding boxes along with their corresponding class labels, which can be extracted as a list of combination \ $\textbf{b}_{pred}=\{(b_1, c_1), (b_2, c_2), \ldots, (b_n, c_n)\}$ by a function $f_{ovd}$, where $b_i$ is the bounding box and $c_i$ is the class label.
Let $q$ denote the input question, $\text{mAP}(\cdot)$ the function calculating mean average precision metric, $\textbf{b}_{gt}$ the list of the combination of ground-truth bounding-boxes and class labels, $L_{gt}$ the number of the ground truth combinations, and $L_{pred}$ the number of the predicted combinations.
The accuracy reward for OVD is defined as:
\begin{align}
    s_{ovd} \quad &= \min(\,1, \frac{L_{gt}}{L_{pred}}) \\
    R^{ovd}_{acc}(q, o) &= s_{ovd}\cdot \text{mAP}(\,\textbf{b}_{pred}, \textbf{b}_{gt})
\end{align}
where $s_{ovd}$ is the penalty factor to a redundant prediction from VLMs, and our experiment shows that this penalty factor is helpful to improve the performance on OVD task. This reward is designated as \textit{odLength} reward.

\noindent\textbf{Format reward}. Format reward of OVD checks whether the response follows the specified format, which requires the model to output a markdown-style JSON response in the \texttt{<answer>} tag~\texttt{(<think>...</think><answer>\allowbreak```json...```</answer>)}, returning 1 or 0 based on compliance.

%% file: sec/5_experiment.tex
\definecolor{darkgreen}{RGB}{0, 150, 0}
\begin{table*}[h!]
\footnotesize
  \begin{center}
  \setlength{\tabcolsep}{11pt}
    \begin{tabular}{c c c c c c c c c c c c} 
    \toprule
       &  & \textbf{Evaluation} &  & \multicolumn{7}{c}{\textbf{Training Steps}}\\
      \textbf{Training method} &  & \textbf{Dataset} & & 0 & 100 & 200 & 300 &  400 & 500 & 600 \\
    \midrule
        SFT & \vline & \multirow{2}{*}{Refcoco$_{val}$}  &  \vline& 88.7  & 88.7 & 88.85 & 88.7 & 88.25 & 88.85 &  88.7 \\
        RL & \vline &   &  \vline & 88.7 & 88.7 & 88.7 & 89.4 & 89.25 & 90 & 90.55 \\
    \noalign{\vskip 0.25ex}
    \hdashline
    \noalign{\vskip 0.25ex}
        SFT & \vline & \multirow{2}{*}{Refcoco+$_{val}$}  &  \vline& 81.95  &82.55& 82.15 &81.85  & 81.9 & 82.3 & 82.25 \\
        RL & \vline &   &  \vline& 81.95 & 82.6 & 81.9 & 82.8 &83.35  &83.6& 84.3 \\
    \noalign{\vskip 0.25ex}
    \hdashline
    \noalign{\vskip 0.25ex}
        SFT & \vline & \multirow{2}{*}{Refcocog$_{val}$}  &  \vline& 86.05  & 85.65 & 85.95 & 85.85 &85.6& 85.95 & 85.95 \\
        
        RL & \vline &   &  \vline& 86.05 & 85.95 &85.05& 85.45 &85.65  & 87.15 & 87.1 \\
    \midrule
        SFT & \vline & \multirow{3}{*}{LISA-Grounding}  &  \vline& 56.51  & 55.91 & 56.51 &55.66  & 55.18 & 55.66 &54.82\\
        RL & \vline &   &  \vline& 56.51 &61.82& 61.27 & 61.64 & 62.6 & 61.88 & 63.14 \\
        \cdashline{5-11}
        \noalign{\vskip 0.25ex}
        $\Delta_{RL-SFT}$ & \vline &   &  \vline & 0 & \textcolor{darkgreen}{+5.91} & \textcolor{darkgreen}{+4.76} & \textcolor{darkgreen}{+5.98} & \textcolor{darkgreen}{+7.42} & \textcolor{darkgreen}{+6.22} & \textcolor{darkgreen}{+8.32} \\
    \bottomrule
    \end{tabular}
  \end{center}
  \caption{Performance comparison of SFT and RL on in-domain and out-of-domain evaluation datasets. All results are from Qwen2.5VL-3B-Instruct trained on the training split of Refcoco/+/g. Step 0 represents the results from Qwen2.5VL-3B-Instruct itself. $\Delta_{RL-SFT}$ denotes the improved value of the RL model compared to the SFT model.}
  \label{table:rec results}
  \vspace{-6mm}
\end{table*}

\section{Experiments}
\subsection{Implementation details}
\noindent\textbf{Selected VLM.}
We employ Qwen2.5VL-3B-Instruct as our base model due to its strong potential performance on vision-language understanding that is expected to be exploited through reinforcement learning, and we also introduce Qwen2.5VL-7B-Instruct and 32B in some experiments to investigate the model size impact.

\noindent\textbf{Hyper-parameters setup.}
When training REC with RL, we adopt the default GRPO parameter settings, setting $N$ to 8, temperature to 0.9, number of iterations to 1, and the KL divergence ratio (i.e., $\beta$) to 0.04. We train the model for 2 epochs, using a learning rate of 1e-6 for both RL and SFT. For OVD, we set only $\beta$ to 0, keeping all other parameters identical.

\tcbset{
  myboxstyle/.style={
    colback=gray!20,     
    colframe=black!70,    
    coltitle=white,       
    fonttitle=\bfseries,  
     fontupper=\itshape,
    boxrule=0.8mm,       
    arc=1mm,               
    boxsep=1mm,             
    left=1mm,              
    right=1mm,              
    top=1mm,               
    bottom=1mm,             
    toptitle=0mm,           
    bottomtitle=0mm,     
    enhanced,                  
  }
}

\noindent\textbf{Prompt template.}
\begin{tcolorbox}[
    myboxstyle, 
    title=Problem Template of REC  
]
Please provide the bounding box coordinates of the region this sentence describes: \{\texttt{query}\}.
\end{tcolorbox}
\begin{tcolorbox}[
    myboxstyle, 
    title=Problem Template of OVD  
]
Please carefully check the image and detect the following objects: \{\texttt{target list}\}. Output each detected target's bbox coordinates in JSON format. The format of the bbox coordinates is:

```json

[{"bbox\_2d": [x1, y1, x2, y2], "label": "target name"}, {"bbox\_2d": [x1, y1, x2, y2], "label": "target name"}]

```.

If there are no such targets in the image, simply respond with None.
\end{tcolorbox}
\begin{tcolorbox}[
    myboxstyle, 
    title=Thinking Prompt  
]
\texttt{\{problem\}} Output the thinking process in $<$think$>$ $<$/think$>$ and final answer in $<$answer$>$ $<$/answer$>$ tags.
\end{tcolorbox}

\begin{figure}[t!]
    \centering
    \includegraphics[width=8cm]{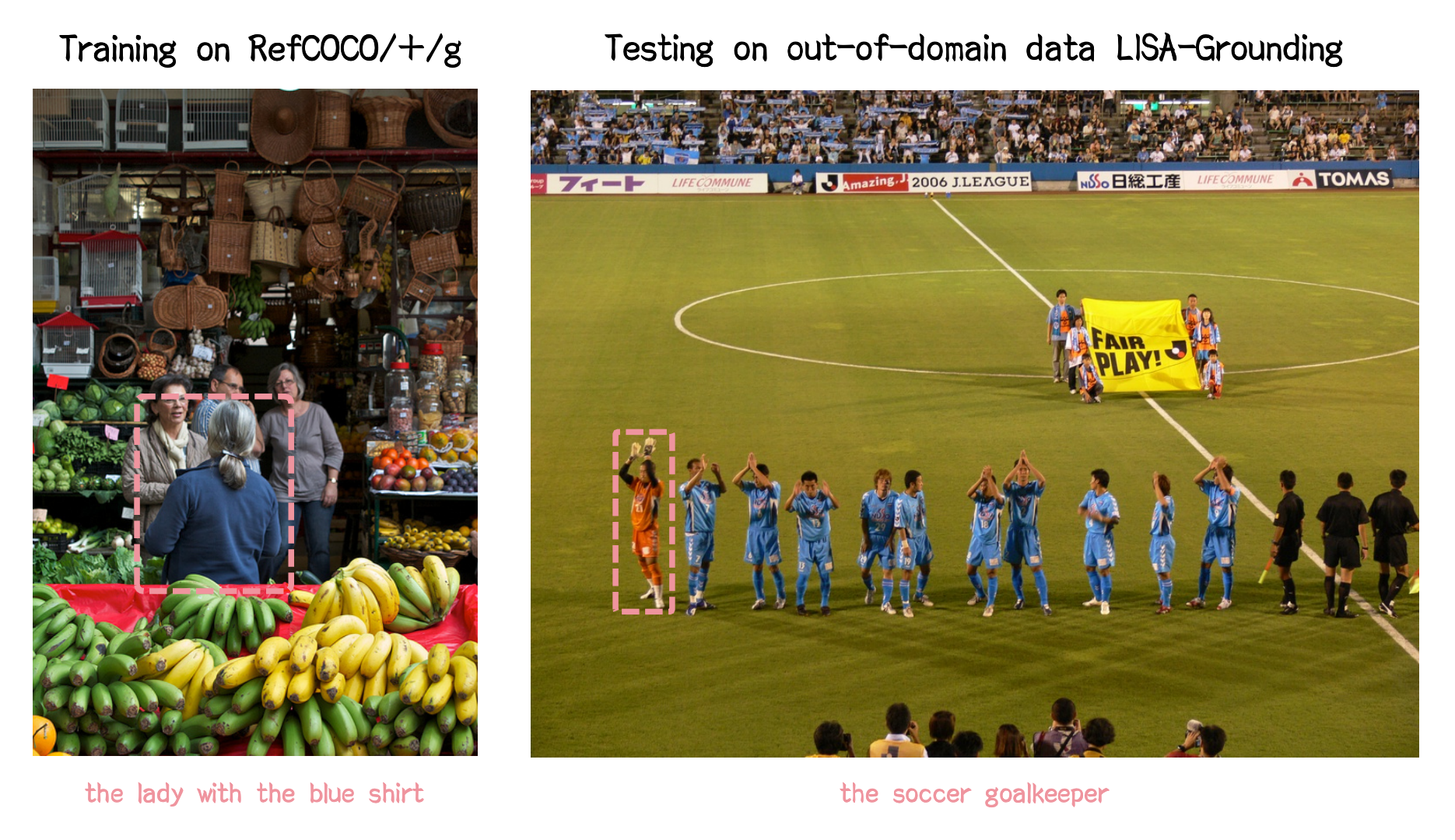}
    \caption{Difference between in-domain and out-of-domain dataset for REC task. In-domain data only describes the spatial or appearance attribute information for the object, while out-of-domain data require the model to use the open-world knowledge to recognize the role of \textit{soccer goalkeeper} and then locate it.}
    \label{fig:dataset}
    \vspace{-3mm}
\end{figure}

\begin{figure}[t!]
    \centering
    \includegraphics[width=7cm]{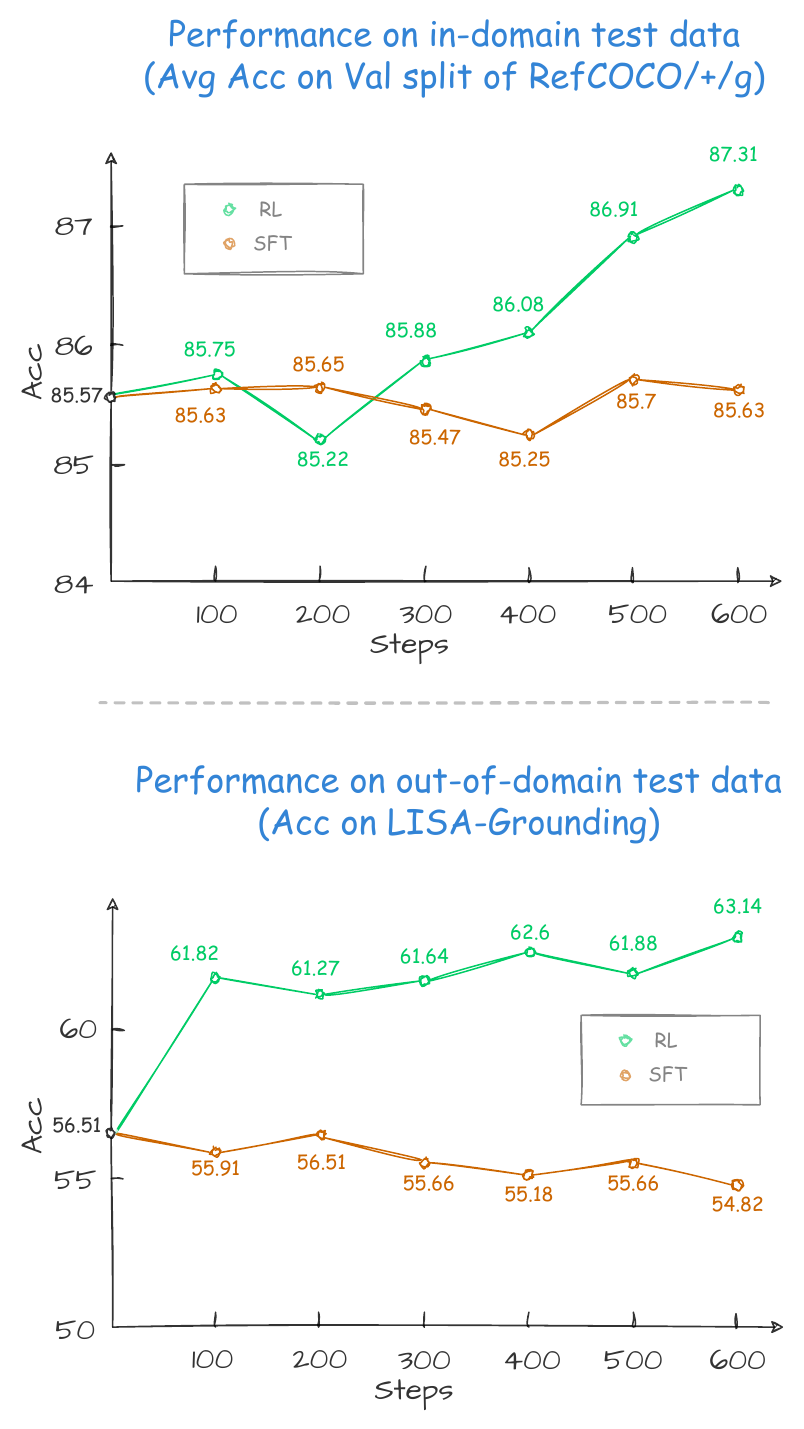}
    \caption{Performance comparison between the SFT and RL models. The RL model shows significantly better generalization on the out-of-domain evaluation dataset compared to the SFT model.}
    \label{fig:performance rec}
    \vspace{-2mm}
\end{figure}

\subsection{Main results}
\subsubsection{Referring expression comprehension}
\noindent\textbf{Training dataset.}
We use the training splits of Refcoco/+/g~\cite{mao2016generation,yu2016modeling} as our training data. These are the most widely used datasets for the REC task and primarily contain descriptions of objects based on spatial or appearance attributes without involving explicit reasoning information. Our objective is to investigate whether a model trained on this non-reasoning dataset can generalize reasoning capabilities acquired through the RL process to more challenging evaluation scenarios.

\noindent\textbf{Evaluation dataset.} We select the val split of Refcoco/+/g~\cite{mao2016generation,yu2016modeling} for in-domain evaluation and test split of LISA-Grounding~\cite{lai2024lisa} for out-of-domain evaluation. 
LISA-Grounding is a reasoning-intensive dataset that requires models to perform fine-grained visual perception, precise understanding of referring expressions, and relational reasoning among objects to correctly localize the target bounding box.
An example of the difference between the two datasets is shown in Figure~\ref{fig:dataset}. 
The evaluation on LISA-Grounding serves as a crucial test of the model's ability to generalize its reasoning skills, acquired from less reasoning-demanding in-domain datasets, to a significantly more challenging out-of-domain scenario.

\noindent\textbf{Results.} 
Table~\ref{table:rec results} shows the performance of the SFT and RL models in four datasets, with a corresponding visualization provided in Figure~\ref{fig:performance rec} for a clearer comparison. In the in-domain test data, the SFT model shows limited improvement over the base model~(i.e., steps 0), regardless of the number of training steps, while the RL model consistently achieves steady performance gains (top of Figure~\ref{fig:performance rec}). More critically, on the out-of-domain test data, the SFT model suffers from a slight performance degradation as training progresses. In contrast, the RL model effectively generalizes its reasoning ability to the out-of-domain setting, maintaining stable and superior performance (bottom of Figure~\ref{fig:performance rec}). These results clearly demonstrate the advantage of reinforcement learning in improving the generalization of VLMs in challenging scenarios that require intense reasoning.

\begin{table}[t!]
\footnotesize
  \begin{center}
  \setlength{\tabcolsep}{4pt}
    \begin{tabular}{l c c c c} 
    \toprule
       &    & \multicolumn{3}{c}{\textbf{COCO}$_{filtered}$}\\
      \textbf{Model} &  & \quad\quad mAP\quad\quad\quad  & GP (IoU=0.5) & GR (IoU=0.5) \\
    \midrule
        Base 3B & \vline & 14.2 & 56.06 & 33.79 \\
        Base 7B & \vline & 14.4 &	54.73 &	33.36 \\
        SFT Model 3B & \vline & 18.5 &	53.15 &	39.4 \\
        RL Model 3B & \vline & \textbf{21.1} &	\textbf{67.34} & \textbf{43.84} \\
    \bottomrule
    \end{tabular}
  \end{center}
  \caption{Results of OVD task on COCO$_{filtered}$. Base 3B denotes the Qwen2.5VL-3B-Instruct and Base 7B denotes the 7B model. GP and GR represents \textit{Greedy Precision} and \textit{Greedy Recall}, respectively.}  
  \label{table:coco results}
  \vspace{-4mm}
\end{table}

\begin{table*}[t!]
\footnotesize
  \begin{center}
  \setlength{\tabcolsep}{5pt}
    \begin{tabular}{l c c c c c c c c c c c} 
    \toprule
      \textbf{Model} &  & \textbf{Celebrity} & \textbf{Logo} & \textbf{Landmark} & \textbf{Color} & \textbf{Material} & \textbf{Position} & \textbf{Relationship} & \textbf{Negation} &  & \textbf{Overall NMS-AP} \\
    \midrule
    \textbf{Specialized OVD Model} & \vline &  &  &  &  &  & &  &  & \vline &  \\
    Grounding-DINO~\cite{liu2024grounding} & \vline & 0.7 & 10.3 & 15.1 & 9.4 & 9.0 & 67.5 & 10.7 & 52.5  & \vline &  25.30 \\
    OmDet~\cite{zhao2024omdet} & \vline & 1.8 & 6.1 & 26.3 & 22.9 & 16.3 & 21.2 & 41.98 & 35.1 & \vline & 25.86 \\
    \midrule
     \textbf{VLM} & \vline &  &  &  &  &  & &  &  & \vline &  \\
      Base 3B & \vline & 13.2 & 26.5 & 21.3 & 2.9 & \textbf{11.6} & \textbf{47.9} & 13.1 & 38.7 & \vline &  25.46\\
      Base 7B & \vline &  48.4 & 35.8 & 44.6 & 3.0 & 10.5 & 40.5 & 16.2 & \textbf{39} & \vline & 29.08 \\
      SFT Model 3B & \vline & 50.4 & \textbf{34.9} & \textbf{50.7} & 4.3 & 7.6 & 33.7 & 13.1 & 34.4  & \vline & 26.50\\
      RL Model 3B & \vline & \textbf{55.0} & 34.6 & 47.9 & \textbf{4.5} & 9.7 & 42.9 & \textbf{21.5} & 37.7  & \vline & \textbf{31.01} \\
      \noalign{\vskip 0.25ex}
    \hdashline
    \noalign{\vskip 0.25ex}
      $\Delta_{RL-SFT}$   & \vline &  \textcolor{darkgreen}{+4.6} & \textcolor{red}{-0.3} & \textcolor{red}{-2.8} & \textcolor{darkgreen}{+0.2} & \textcolor{darkgreen}{+2.1} & \textcolor{darkgreen}{+9.2} & \textcolor{darkgreen}{+8.4} & \textcolor{darkgreen}{+3.3} & \vline & \textcolor{darkgreen}{+4.51} \\
    \bottomrule
    \end{tabular}
  \end{center}
  \caption{Results of OVD task on OVDEval. Base denotes the Qwen2.5VL-3B-Instruct, and Base 7B denotes the 7B model. $\Delta_{RL-SFT}$ denotes the improved value of the RL model compared to the SFT model. We also list the performance of OmDet, the current state-of-the-art in specialized open-vocabulary detection, for the comprehensive comparison.}
  \label{table:ovdeval results}
  \vspace{-3mm}
\end{table*}

\subsubsection{Open-vocabulary object detection}
\noindent\textbf{Training dataset.}
We use Description Detection Dataset~(D$^3$)\cite{xie2023described} as our training data, which provides several unique advantages for training object detection models:~(1)~complete annotation coverage, (2)~unrestricted language descriptions, (3)~instance-level annotations, and (4)~absence expression support. 
During training, we randomly introduce 1$\sim$3 descriptions from other training samples as negative expressions.

\noindent\textbf{Evaluation dataset.} We select COCO$_{filtered}$ and OVDEval~\cite{yao2023evaluate} for evaluation. COCO$_{filtered}$ is created from the COCO~\cite{lin2014microsoft} dataset's \texttt{instances$\_$val2017.json} file. Since VLMs generally struggle at recall in OD tasks~(see \cite{jiang2024chatrextamingmultimodalllm} for details), we filter out categories with more than 10 annotation boxes, ensuring that only categories with fewer boxes are included.
OVDEval is utilized to evaluate the model's capabilities. This is a comprehensive benchmark specifically designed for open-vocabulary detection, which systematically evaluates models across six key linguistic aspects~\footnote{Including \textit{Object detection, Proper noun recognition (celebrities, logos, landmarks), Attribute detection, Position understanding, Relationship comprehension }, and \textit{Negation handling}}. It further introduces hard negative samples to assess robustness and a novel NMS-AP metric to address the ``Inflated AP Problem" common issues, providing a more accurate OVD assessment. All output boxes generated from VLM are assigned a \textbf{confidence score of 1} when calculating AP.
During COCO evaluation, \texttt{\{target list\}} is consistently set as all COCO 80 categories. For OVDEval evaluation, we keep the official evaluation setting.

\noindent\textbf{Results.}
Table~\ref{table:coco results} shows the performance on COCO$_{filtered}$. The RL-trained model demonstrated substantial improvements over the SFT model, with a 2.6 percentage point increase in mAP (21.1\% vs 18.5\%), 4.42 points higher in Greedy Precision (57.57\% vs 53.15\%), and 4.33 points better in Greedy Recall (43.73\% vs 39.4\%). These consistent improvements across all metrics demonstrate RL's superior generalization capability.

On the more challenging and comprehensive benchmark OVDEval, from Table~\ref{table:ovdeval results}, it is observed that the RL model demonstrated superior generalization by outperforming SFT in 7 out of 9 detection categories. Most notably, it shows significant improvements in complex tasks requiring deeper understanding: \textit{Position detection}~(+9.2 points), \textit{Relationship detection}~(+8.4 points), and \textit{Negation handling}~(+3.3 points). Moreover, although SFT shows strong performance in specific categories like \textit{Celebrity}, \textit{Logo}, and \textit{Landmark} detection, RL demonstrates more balanced improvements across different visual tasks, suggesting better overall generalization of visual understanding.

The results demonstrate that while SFT can be effective for certain specific tasks, RL provides more comprehensive improvements. The 4.51 point improvement in average nms-ap (31.01 vs 26.50) indicates RL's superior ability to learn generalizable features.

\begin{table*}[t!]
\footnotesize
  \begin{center}
  \setlength{\tabcolsep}{8pt}
    \begin{tabular}{l c c c c c c c c c c c c c} 
    \toprule
     & \multicolumn{3}{c}{COCO$_{filtered}$} &  &\multicolumn{9}{c}{OVDEval} \\
      \textbf{Reward} & \textbf{mAP} & \textbf{GP} & \textbf{GR} &  &\textbf{Cel} & \textbf{Logo} & \textbf{Land} & \textbf{Color} & \textbf{Mat} & \textbf{Pos} & \textbf{Rel} & \textbf{Neg} &  \textbf{Overall} \\
    \midrule
      $\text{AP}_{50}$ & 11.4 &	45.33 &	31.85 &\vline &7.7 & 13.8 &17.3 & 2.2 & 3.5 & 39.3 & 17.0 & 35.2  &  21.46\\
      $\text{mAP}$ & 11.8 &	46.02 &	33.34 &\vline & 10.5 & 15.2 & 18.7 & 2.3 &4.1 & 39.6 & 15.9 & 34.9  & 21.68\\
      $s_{ovd}\cdot\text{mAP}$ & \textbf{21.1} &	\textbf{67.34} & \textbf{43.84}  &\vline & \textbf{55.0} & \textbf{34.6} & \textbf{47.9} & \textbf{4.5} & \textbf{9.7} & \textbf{42.9} & \textbf{21.5} & \textbf{37.7}  &  \textbf{31.01}\\
    \bottomrule
    \end{tabular}
  \end{center}
  \caption{Performance comparison across \textit{AP}$_{50}$ reward, \textit{mAP} reward and \textit{odLength} reward. All results are obtained by the RL model trained from Qwen2.5VL-3B-Instruct. GP: Greedy Precision; GR: Greedy Recall; Cel: Celebrity; Land: Landmark; Mat: Material; Pos: Position; Rel: Relationship; Neg: Negation .}
  \label{table:odLength}
  \vspace{-3mm}
\end{table*}

\noindent\textbf{Comparison with SoTA OD: OmDet.} 
OmDet~\cite{zhao2024omdet} represents the current state-of-the-art in specialized open-vocabulary detection. However, our VLM-R1 model demonstrates that VLMs can outperform specialized architectures in several key aspects.

The performance gap between RL model and OmDet reveals interesting insights about the \textbf{strengths and limitations} of different approaches:
\begin{itemize}
    \item \textit{World Knowledge and Entity Recognition:} In celebrity detection, VLM-R1 achieves 55.0 nms-ap compared to OmDet's 1.8. This dramatic difference ($>$50 points) demonstrates the value of VLMs' pre-trained world knowledge, and similar patterns appear in logo and landmark detection, where semantic understanding is crucial.
    \item \textit{Fine-grained Detection:} We note that the attribute category in OVDEval contains a lot of small objects. In these small-object detection scenarios, OmDet shows a stronger performance gap~(color: 22.9 vs 4.5). This suggests specialized architectures excel at fine-grained, local feature detection.
\end{itemize}
These comparisons suggest a promising future direction: combining the complementary strengths of both approaches. Specialized OD architectures excel at fine-grained detection and high-recall scenarios, while VLMs bring rich world knowledge. Future research could focus on creating hybrid architectures that leverage both the precise localization abilities of dedicated OD models and the semantic understanding of VLMs.

\subsection{Ablations \& Extended experiments}

\subsubsection{Investigation about ``reward hacking"}

\noindent\textbf{What is reward hacking?}
Reward hacking~\cite{amodei2016concrete} in reinforcement learning refers to a phenomenon where an agent exploits loopholes in the reward function to achieve high reward without truly fulfilling the intended task. This occurs when the reward function is misaligned with the designer’s actual goals, leading the agent to adopt unintended or shortcut behaviors. For instance, in a maze navigation task where the agent receives +1 for each step and +100 for exiting the maze, the agent may learn to walk in circles indefinitely to accumulate step rewards rather than solving the maze. Such behavior technically maximizes reward but fails to meet the task’s true objective. Several literature~\cite {denison2024sycophancy,pan2024feedback,liu2023llms,pan2024spontaneous,wang2023large,wen2024language} also investigate this phenomenon in large language model research.

\noindent\textbf{Reward hacking in OVD task.}
Table~\ref{table:odLength} exhibits the superior performance of our proposed \textit{odLength} compared to the native \textit{AP}$_{50}$ and \textit{mAP} reward. Upon closer examination, we identify key limitations of the native \textit{AP}${50}$ and \textit{mAP} reward functions. Specifically, we observe that when computing AP values using the official COCO evaluation API, categories not present in the ground truth for a given image are excluded from the evaluation. Given our prompt design, which consistently includes all positive and several negative categories, the model is incentivized to predict all categories in order to maximize its reward---\textbf{an instance of reward hacking}. This behavior negatively impacts precision when evaluated on the full dataset, where all COCO 80 categories are present, and no category will be excluded. 
In contrast, our \textit{odLength} reward addresses this issue by introducing an additional penalty term for redundant predictions. This encourages the model to align the number of predicted objects with the ground truth, thereby promoting more precise and faithful outputs from VLMs.


\begin{figure}[t!]
    \centering
    \includegraphics[width=8.5cm]{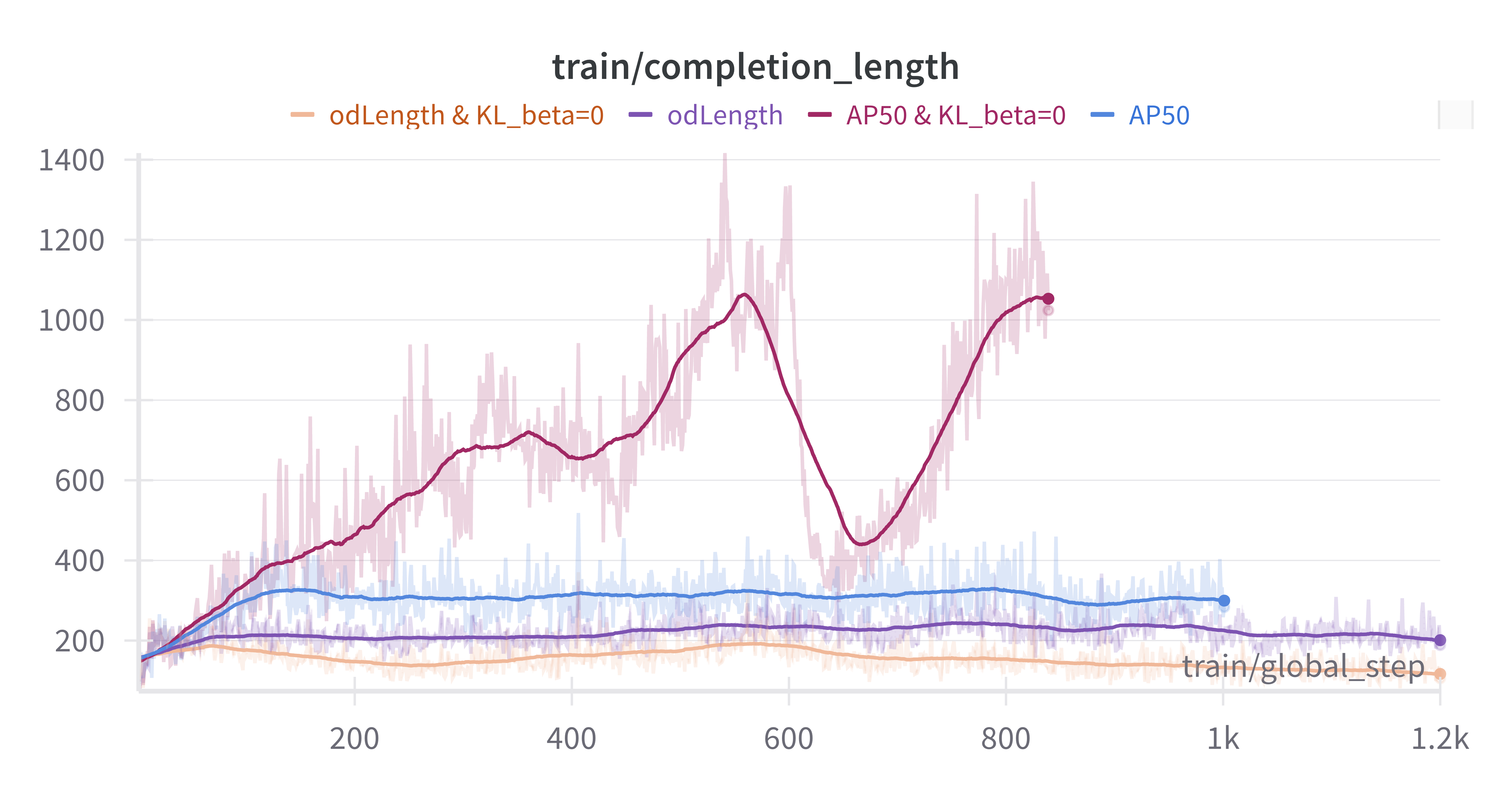}
    \caption{Visualization of the completion length across different reward settings on OVD task. It is observed that the model always generates the overlong completion with the native \textit{AP} reward, indicating the redundant predicted objects.}
    \label{fig:length}
    \vspace{-2mm}
\end{figure}

\begin{figure*}[ht]
    \centering
    \includegraphics[width=16cm]{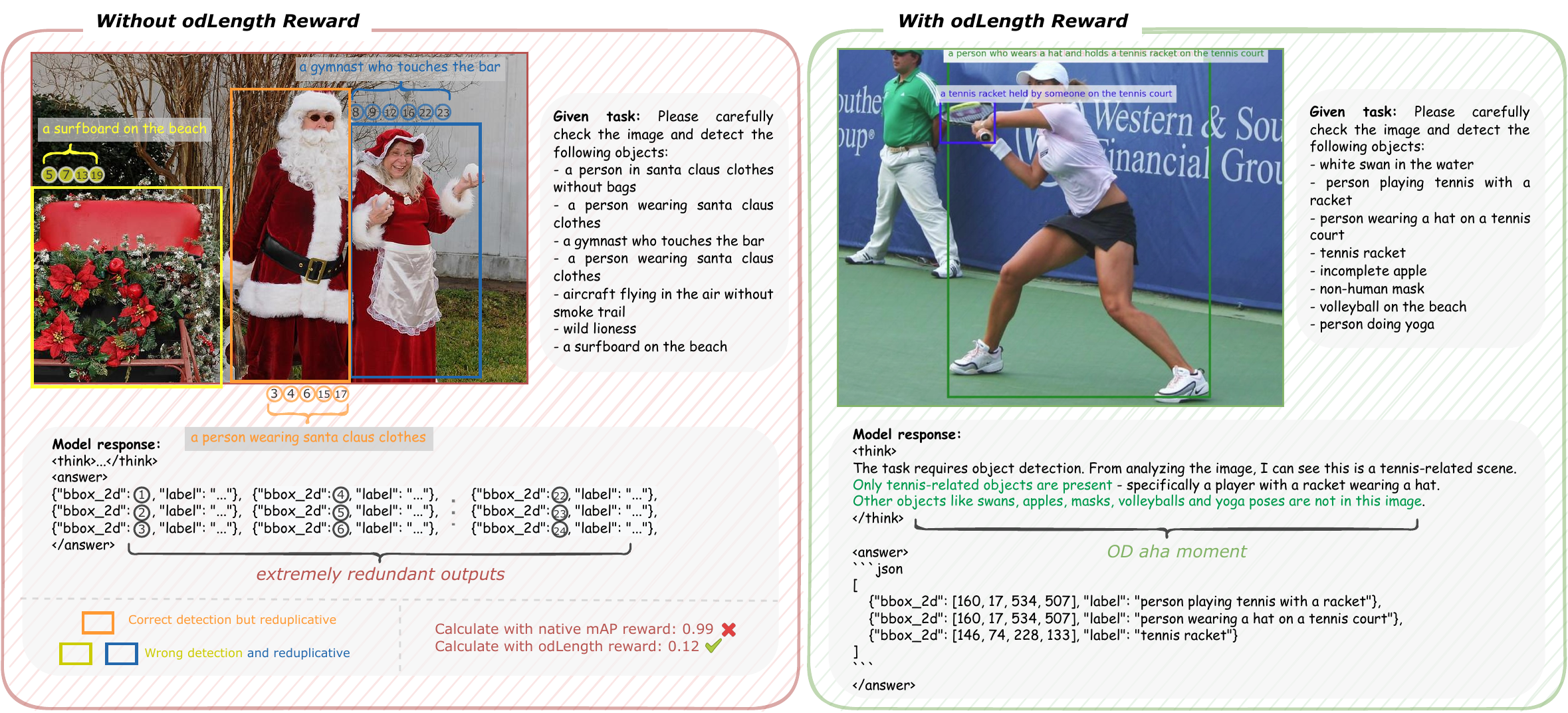}
     \caption{Comparison of cases with and without the proposed \textit{odLength} reward. \textbf{Left}: Without \textit{odLength}, the model generates redundant and duplicated boxes, yet still receives a high reward from native \textit{mAP}. Each circle denotes a predicted bounding box, and circles of the same color indicate bounding boxes with identical coordinates. \textbf{Right}: With \textit{odLength}, the model exhibits an “OD aha moment”, first reasoning about object presence before producing accurate bounding boxes.}
    \label{fig:odLength}
\end{figure*}

\begin{table*}[t!]
\footnotesize
  \begin{center}
  \setlength{\tabcolsep}{7pt}
    \begin{tabular}{l c c c c c c c c c c c c c} 
    \toprule
     & \multicolumn{3}{c}{COCO$_{filtered}$} &  &\multicolumn{9}{c}{OVDEval} \\
      \textbf{Model} & \textbf{mAP} & \textbf{GP} & \textbf{GR} &  &\textbf{Cel} & \textbf{Logo} & \textbf{Land} & \textbf{Color} & \textbf{Mat} & \textbf{Pos} & \textbf{Rel} & \textbf{Neg} &  \textbf{Overall} \\
    \midrule
     Qwen2.5VL-3B-Instruct & 14.2 & 56.06 & 33.79 &\vline & 13.2 & 26.5 & 21.3 & 2.9 & \textbf{11.6} & \textbf{47.9} & 13.1 & \textbf{38.7}  &  25.46\\
      \quad w/ COCO training &12.6 & 48.43 & 31.86 &\vline & 9.9 & 18.3 & 26.6 & 2.5 & 6.2 &45.3 & 18.1 & 36.4 &  24.48\\
      \quad w/ D$^3$ training & \textbf{21.1} & \textbf{67.34} & \textbf{43.84} & \vline & \textbf{55.0} & \textbf{34.6} & \textbf{47.9} & \textbf{4.5} & 9.7 & 42.9 & \textbf{21.5} & 37.7   & \textbf{31.01}\\
    \bottomrule
    \end{tabular}
  \end{center}
  \caption{Performance comparison of models trained on different training data. GP: Greedy Precision; GR: Greedy Recall; Cel: Celebrity; Land: Landmark; Mat: Material; Pos: Position; Rel: Relationship; Neg: Negation .}
  \label{table:performance training data}
\end{table*}

\noindent\textbf{Visualization of the completion length.}
Figure~\ref{fig:length} illustrates the variation in output sequence lengths across different reward settings. Notably, models trained with the native \textit{AP}$_{50}$ reward---particularly those without KL regularization---exhibit a dramatic increase in output length over the course of training. This trend indicates the presence of severe reward hacking, where the model is incentivized to enumerate an excessive number of object categories in order to maximize the reward, leading to highly redundant outputs. In contrast, models trained with our proposed \textit{odLength} reward maintain stable and significantly shorter outputs, effectively suppressing unnecessary predictions. 

\noindent\textbf{OD aha moment.}
Figure~\ref{fig:odLength} illustrates the cases with and without the proposed \textit{odLength} reward. Without the \textit{odLength} reward, the VLM produces extremely redundant outputs, including both correct-but-duplicated and incorrect-but-duplicated detections. Despite the poor quality of the detection results, the native \textit{mAP} still assigns a relatively high reward, revealing its susceptibility to reward hacking. However, with our proposed \textit{odLength} reward,  the VLM is incentivized to precisely locate every object, demonstrating an emergent reasoning behavior, which we refer to as the ``\textbf{OD aha moment}". Faced with complex detection tasks involving multiple potential targets (including hard negatives), the model spontaneously adopts a two-step strategy: it first identifies which objects are truly present through an explicit “thinking” step, and then proceeds with accurate bounding box prediction.


\begin{table*}[t!]
\footnotesize
  \begin{center}
  \setlength{\tabcolsep}{7pt}
    \begin{tabular}{l c c c c c c c c c c c c c} 
    \toprule
     & \multicolumn{3}{c}{COCO$_{filtered}$} &  &\multicolumn{9}{c}{OVDEval} \\
      \textbf{Model} & \textbf{mAP} & \textbf{GP} & \textbf{GR} &  &\textbf{Cel} & \textbf{Logo} & \textbf{Land} & \textbf{Color} & \textbf{Mat} & \textbf{Pos} & \textbf{Rel} & \textbf{Neg} &  \textbf{Overall} \\
    \midrule
      Qwen2.5VL-3B-Instruct& 14.2 & 56.06 & 33.79 &\vline & 13.2 & 26.5 & 21.3 & 2.9 & 11.6 & \textbf{47.9} & 13.1 & 38.7  &  25.46\\
      \quad w/ RL  & 21.1 & 67.34 & 43.84 & \vline & 55.0 & 34.6 & 47.9 & 4.5 & 9.7 & 42.9 & 21.5 & 37.7   & 31.01\\ \hline
      Qwen2.5VL-7B-Instruct & 14.4 & 54.73 & 33.36 &\vline & 48.4 & 35.8 & 44.6 & 3.0 & 10.5 & 40.5 & 16.2 & 39.0  & 29.08\\
      \quad w/ RL  &21.9 & \textbf{74.46} & 41.2 &\vline & 57.1 & \textbf{38.3} & \textbf{49.4} & \textbf{7.8} & 14.7 & 39.4 & 20.1 & 43.1 &  32.42\\ 
      \hline
      Qwen2.5VL-32B-Instruct & 18.6 & 57.26 & 47.58  &\vline & 57.7 & 32.5 & 46.7 & 4.4 & 13.6 & 41.7 & 20.6 & 47.0  &  32.79\\
      \quad w/ RL  & \textbf{23.0} &74.04& \textbf{48.67} &\vline & \textbf{57.8} & 35.8 & 48.3 & \textbf{7.8} & \textbf{19.1} & 44.5 & \textbf{27.0} & \textbf{51.7}  &  \textbf{36.79}\\
    \bottomrule
    \end{tabular}
  \end{center}
  \caption{Performance comparison across models with different sizes and their corresponding RL models. GP: Greedy Precision; GR: Greedy Recall; Cel: Celebrity; Land: Landmark; Mat: Material; Pos: Position; Rel: Relationship; Neg: Negation.}
  \label{table:performance model size}
  \vspace{-2mm}
\end{table*}



\subsubsection{Selection of training data}
Table~\ref{table:performance training data} presents a performance comparison between models trained on COCO and D$^3$ datasets. Notably, the model trained on D$^3$ significantly outperforms the one trained on COCO—even on the in-domain COCO$_{filtered}$ evaluation set, which aligns with the distribution of the COCO training data. One key difference lies in the semantic complexity of the training queries: COCO categories are typically simple, often consisting of single-word labels (e.g., \textit{person, car}), whereas D$^3$ queries are semantically richer, typically formulated as full, meaning-intensive sentences (see Figure~\ref{fig:odLength} for examples).
We hypothesize that this difference in semantic richness plays a pivotal role in the observed performance gap. In the context of reinforcement learning, challenging and semantically complex data is essential for encouraging the model to develop more robust reasoning chains, ultimately leading to superior task performance.

\subsubsection{RL effects across model scales}\label{sec:model size impact}
Table~\ref{table:performance model size} presents the performance comparison between models of different sizes and their corresponding RL-enhanced versions. Several noteworthy observations emerge:
\begin{itemize}
    \item The Relation sub-task, which requires reasoning ability, shows a substantial performance boost after applying RL across all model sizes~(13.1$\rightarrow$21.5, 16.2$\rightarrow$20.1, 20.6$\rightarrow$~27.0), indicating that RL could exploit the superior reasoning capabilities of VLMs.
    \item Another reasoning-intensive sub-task, Negation, both 7B and 32B RL model achieve improvement~(39.0$\rightarrow$43.1, 47.0$\rightarrow$51.7), whereas 3B model suffers from slight performance degradation~(38.7$\rightarrow$~37.7). We posit that this discrepancy arises from the inherent capacity of the base models. As demonstrated by \cite{liu2025understanding}, reinforcement learning predominantly serves to reinforce correct reasoning patterns rather than infuse novel knowledge. Given the greater capacity of the 7B and 32B base models, it is plausible that reinforcement learning more effectively harnesses their latent reasoning abilities.
    \item In the context of the Color sub-task, 7B and 32B RL model exhibit more performance gains than 3B model~(2.9~$\rightarrow$4.5 vs. 3.0$\rightarrow$7.8, 4.4$\rightarrow$7.8). Given that the Color sub-task in OVDEval primarily involves small objects, this comparison highlights the superior visual perception capabilities about fine-grained visual details of the large VLMs.
    \item On the COCO$_{filtered}$ subset, models across all sizes demonstrate greater gains in Greedy Precision relative to Greedy Recall. This discrepancy aligns with the design of the \textit{odLength} reward, which explicitly penalizes redundant bounding box predictions. While this amendment improves precision by discouraging over-prediction, it can lead to a slight reduction in recall due to the model’s increased conservativeness when outputting predictions.
    \item Larger models generally perform slightly better.
\end{itemize}

%% file: sec/6_discussion.tex
\section{Discussion}

\subsection{Reinforcement Learning vs. Supervised Fine-Tuning}
In the context of referring expression comprehension, in addition to achieving steady performance gains on in-domain tasks, the RL model generalizes the reasoning patterns acquired from the non-reasoning training data to the out-of-domain settings that require a more nuanced understanding and complex reasoning. This suggests that RL not only optimizes for performance on seen scenarios, but also encourages models to develop transferable capabilities applicable to more challenging, unseen tasks.

Furthermore, in the open-vocabulary object detection experiments, RL models outperform their supervised SFT counterparts in most subtasks on the complex OVDEval benchmark, particularly achieving substantial gains in some challenging subtasks. Moreover, as discussed in\ref{sec:model size impact}, models of nearly all sizes benefit from RL in these reasoning-focused tasks, further validating the generalization advantage of this training paradigm.

These findings strongly support the conclusion proposed by~\cite{chu2025sft}: \textit{``SFT Memorizes, RL Generalizes"}. Our results reinforce the effectiveness of RL in enhancing the generalization capabilities of VLMs, especially in scenarios that require reasoning patterns.

\subsection{Preventing Reward Hacking via Reward Engineering}
In this report, we reveal the phenomenon of reward hacking on OVD tasks when using the native \textit{mAP} reward and demonstrate the effectiveness of our proposed \textit{odLength} reward in mitigating this issue. As illustrated in Figure~\ref{fig:odLength}, the poorly designed reward function incentivizes the model to generate excessive and indiscriminate predictions in pursuit of higher reward values. This behavior results in degraded performance on evaluation benchmarks. By contrast, incorporating the \textit{odLength} reward significantly suppresses such redundant outputs, leading to improved alignment between reward signals and evaluation metrics, and more importantly, emerging \textit{``OD aha moment"}, first reasoning about object presence before producing accurate bounding boxes.

These results highlight the importance of careful reward design in reinforcement learning pipelines, particularly for complex tasks where naively defined objectives may fail to capture desired model behavior.

\subsection{Role of Data in Reasoning and Generalization}
Our findings highlight the pivotal role of training data in shaping model performance. We observe that complex and challenging training samples can effectively elicit reasoning behaviors in VLMs, consistent with the observations in~\cite{meng2025mm}. Conversely, low-quality or over-simple data may hinder learning and even negatively impact generalization~(Table~\ref{table:performance training data}). These insights emphasize the necessity of careful training data selection.

Equally important is the choice of evaluation data. Comprehensive and appropriately challenging benchmarks are essential for accurately assessing a model’s reasoning and perception capabilities. In this study, we select LISA-Grounding and OVDEval as our evaluation datasets, as they are both designed to probe complex semantic understanding and generalization in complicated, real-world scenarios. Together, our results reinforce the importance of high-quality training and evaluation data to advance the capabilities of VLMs.

\subsection{From Simple to Complex: Adapting RL for OVD}
In this report, we explore the feasibility of applying the R1-style reinforcement learning framework to two structurally similar tasks: referring expression comprehension (REC) and open-vocabulary object detection (OVD), both of which require the model to output bounding boxes based on textual descriptions. Despite their surface similarity, our comparative analysis reveals that additional optimization is essential to successfully apply RL to the more complex OVD task.

First, while a naive reward function suffices for REC, it fails to yield effective training in OVD due to reward hacking, necessitating the design of a more robust, customized reward---such as our proposed \textit{odLength}. Second, although models trained on relatively simple in-domain datasets~(i.e., RefCOCO) generalize well in the REC setting, the same approach does not transfer effectively to OVD. To address this, we carefully choose a more appropriate training dataset for OVD~(i.e., D$^3$), thus achieving a superior result.

These findings underscore the need for task-specific optimizations when applying RL to more complex scenarios.

%% file: main.bbl
\begin{thebibliography}{61}
\providecommand{\natexlab}[1]{#1}
\providecommand{\url}[1]{\texttt{#1}}
\expandafter\ifx\csname urlstyle\endcsname\relax
  \providecommand{\doi}[1]{doi: #1}\else
  \providecommand{\doi}{doi: \begingroup \urlstyle{rm}\Url}\fi

\bibitem[zhe(2025)]{zheng2025easyr1}
Easyr1: An efficient, scalable, multi-modality rl training framework.
\newblock \url{https://github.com/hiyouga/EasyR1}, 2025.

\bibitem[Achiam et~al.(2023)Achiam, Adler, Agarwal, Ahmad, Akkaya, Aleman, Almeida, Altenschmidt, Altman, Anadkat, et~al.]{achiam2023gpt}
Josh Achiam, Steven Adler, Sandhini Agarwal, Lama Ahmad, Ilge Akkaya, Florencia~Leoni Aleman, Diogo Almeida, Janko Altenschmidt, Sam Altman, Shyamal Anadkat, et~al.
\newblock Gpt-4 technical report.
\newblock \emph{arXiv preprint arXiv:2303.08774}, 2023.

\bibitem[Ahmadian et~al.(2024)Ahmadian, Cremer, Gall{\'e}, Fadaee, Kreutzer, Pietquin, {\"U}st{\"u}n, and Hooker]{ahmadian2024back}
Arash Ahmadian, Chris Cremer, Matthias Gall{\'e}, Marzieh Fadaee, Julia Kreutzer, Olivier Pietquin, Ahmet {\"U}st{\"u}n, and Sara Hooker.
\newblock Back to basics: Revisiting reinforce style optimization for learning from human feedback in llms.
\newblock \emph{arXiv preprint arXiv:2402.14740}, 2024.

\bibitem[Alayrac et~al.(2022)Alayrac, Donahue, Luc, Miech, Barr, Hasson, Lenc, Mensch, Millican, Reynolds, et~al.]{alayrac2022flamingo}
Jean-Baptiste Alayrac, Jeff Donahue, Pauline Luc, Antoine Miech, Iain Barr, Yana Hasson, Karel Lenc, Arthur Mensch, Katherine Millican, Malcolm Reynolds, et~al.
\newblock Flamingo: a visual language model for few-shot learning.
\newblock \emph{Advances in neural information processing systems}, 35:\penalty0 23716--23736, 2022.

\bibitem[Amodei et~al.(2016)Amodei, Olah, Steinhardt, Christiano, Schulman, and Man{\'e}]{amodei2016concrete}
Dario Amodei, Chris Olah, Jacob Steinhardt, Paul Christiano, John Schulman, and Dan Man{\'e}.
\newblock Concrete problems in ai safety.
\newblock \emph{arXiv preprint arXiv:1606.06565}, 2016.

\bibitem[Bai et~al.(2025)Bai, Chen, Liu, Wang, Ge, Song, Dang, Wang, Wang, Tang, et~al.]{bai2025qwen2}
Shuai Bai, Keqin Chen, Xuejing Liu, Jialin Wang, Wenbin Ge, Sibo Song, Kai Dang, Peng Wang, Shijie Wang, Jun Tang, et~al.
\newblock Qwen2. 5-vl technical report.
\newblock \emph{arXiv preprint arXiv:2502.13923}, 2025.

\bibitem[Chen et~al.(2024{\natexlab{a}})Chen, Thapa, Chalamala, Athiwaratkun, Song, and Zou]{chen2024dragonfly}
Kezhen Chen, Rahul Thapa, Rahul Chalamala, Ben Athiwaratkun, Shuaiwen~Leon Song, and James Zou.
\newblock Dragonfly: Multi-resolution zoom supercharges large visual-language model.
\newblock \emph{arXiv preprint arXiv:2406.00977}, 2024{\natexlab{a}}.

\bibitem[Chen et~al.(2023)Chen, Li, Dong, Zhang, He, Wang, Zhao, and Lin]{chen2023sharegpt4v}
Lin Chen, Jisong Li, Xiaoyi Dong, Pan Zhang, Conghui He, Jiaqi Wang, Feng Zhao, and Dahua Lin.
\newblock Sharegpt4v: Improving large multi-modal models with better captions.
\newblock \emph{arXiv preprint arXiv:2311.12793}, 2023.

\bibitem[Chen et~al.(2025)Chen, Li, Zhao, Song, and Vinci]{chen2025r1v}
Liang Chen, Lei Li, Haozhe Zhao, Yifan Song, and Vinci.
\newblock R1-v: Reinforcing super generalization ability in vision-language models with less than \$3.
\newblock \url{https://github.com/Deep-Agent/R1-V}, 2025.
\newblock Accessed: 2025-02-02.

\bibitem[Chen et~al.(2024{\natexlab{b}})Chen, Wang, Cao, Liu, Gao, Cui, Zhu, Ye, Tian, Liu, et~al.]{chen2024expanding}
Zhe Chen, Weiyun Wang, Yue Cao, Yangzhou Liu, Zhangwei Gao, Erfei Cui, Jinguo Zhu, Shenglong Ye, Hao Tian, Zhaoyang Liu, et~al.
\newblock Expanding performance boundaries of open-source multimodal models with model, data, and test-time scaling.
\newblock \emph{arXiv preprint arXiv:2412.05271}, 2024{\natexlab{b}}.

\bibitem[Chen et~al.(2024{\natexlab{c}})Chen, Wang, Tian, Ye, Gao, Cui, Tong, Hu, Luo, Ma, et~al.]{chen2024internvl}
Zhe Chen, Weiyun Wang, Hao Tian, Shenglong Ye, Zhangwei Gao, Erfei Cui, Wenwen Tong, Kongzhi Hu, Jiapeng Luo, Zheng Ma, et~al.
\newblock How far are we to gpt-4v? closing the gap to commercial multimodal models with open-source suites.
\newblock \emph{arXiv preprint arXiv:2404.16821}, 2024{\natexlab{c}}.

\bibitem[Chu et~al.(2025)Chu, Zhai, Yang, Tong, Xie, Schuurmans, Le, Levine, and Ma]{chu2025sft}
Tianzhe Chu, Yuexiang Zhai, Jihan Yang, Shengbang Tong, Saining Xie, Dale Schuurmans, Quoc~V Le, Sergey Levine, and Yi Ma.
\newblock Sft memorizes, rl generalizes: A comparative study of foundation model post-training.
\newblock \emph{arXiv preprint arXiv:2501.17161}, 2025.

\bibitem[Dai et~al.(2023)Dai, Li, Li, Tiong, Zhao, Wang, Li, Fung, and Hoi]{instructblip}
Wenliang Dai, Junnan Li, Dongxu Li, Anthony Meng~Huat Tiong, Junqi Zhao, Weisheng Wang, Boyang Li, Pascale Fung, and Steven Hoi.
\newblock Instructblip: Towards general-purpose vision-language models with instruction tuning, 2023.

\bibitem[Deng et~al.(2025)Deng, Zou, Ma, Luo, Cao, and Kang]{deng2025boosting}
Huilin Deng, Ding Zou, Rui Ma, Hongchen Luo, Yang Cao, and Yu Kang.
\newblock Boosting the generalization and reasoning of vision language models with curriculum reinforcement learning.
\newblock \emph{arXiv preprint arXiv:2503.07065}, 2025.

\bibitem[Denison et~al.(2024)Denison, MacDiarmid, Barez, Duvenaud, Kravec, Marks, Schiefer, Soklaski, Tamkin, Kaplan, et~al.]{denison2024sycophancy}
Carson Denison, Monte MacDiarmid, Fazl Barez, David Duvenaud, Shauna Kravec, Samuel Marks, Nicholas Schiefer, Ryan Soklaski, Alex Tamkin, Jared Kaplan, et~al.
\newblock Sycophancy to subterfuge: Investigating reward-tampering in large language models.
\newblock \emph{arXiv preprint arXiv:2406.10162}, 2024.

\bibitem[Face(2025)]{openr1}
Hugging Face.
\newblock Open r1: A fully open reproduction of deepseek-r1, 2025.

\bibitem[Guo et~al.(2025)Guo, Yang, Zhang, Song, Zhang, Xu, Zhu, Ma, Wang, Bi, et~al.]{guo2025deepseek}
Daya Guo, Dejian Yang, Haowei Zhang, Junxiao Song, Ruoyu Zhang, Runxin Xu, Qihao Zhu, Shirong Ma, Peiyi Wang, Xiao Bi, et~al.
\newblock Deepseek-r1: Incentivizing reasoning capability in llms via reinforcement learning.
\newblock \emph{arXiv preprint arXiv:2501.12948}, 2025.

\bibitem[Hu et~al.(2022)Hu, Shen, Wallis, Allen-Zhu, Li, Wang, Wang, Chen, et~al.]{hu2022lora}
Edward~J Hu, Yelong Shen, Phillip Wallis, Zeyuan Allen-Zhu, Yuanzhi Li, Shean Wang, Lu Wang, Weizhu Chen, et~al.
\newblock Lora: Low-rank adaptation of large language models.
\newblock \emph{ICLR}, 1\penalty0 (2):\penalty0 3, 2022.

\bibitem[Huang et~al.(2025)Huang, Jia, Zhai, Cao, Ye, Zhao, Hu, and Lin]{huang2025vision}
Wenxuan Huang, Bohan Jia, Zijie Zhai, Shaosheng Cao, Zheyu Ye, Fei Zhao, Yao Hu, and Shaohui Lin.
\newblock Vision-r1: Incentivizing reasoning capability in multimodal large language models.
\newblock \emph{arXiv preprint arXiv:2503.06749}, 2025.

\bibitem[Jaech et~al.(2024)Jaech, Kalai, Lerer, Richardson, El-Kishky, Low, Helyar, Madry, Beutel, Carney, et~al.]{jaech2024openai}
Aaron Jaech, Adam Kalai, Adam Lerer, Adam Richardson, Ahmed El-Kishky, Aiden Low, Alec Helyar, Aleksander Madry, Alex Beutel, Alex Carney, et~al.
\newblock Openai o1 system card.
\newblock \emph{arXiv preprint arXiv:2412.16720}, 2024.

\bibitem[Jiang et~al.(2024)Jiang, luo, Yang, Xiong, Chen, Zeng, Ren, and Zhang]{jiang2024chatrextamingmultimodalllm}
Qing Jiang, Gen luo, Yuqin Yang, Yuda Xiong, Yihao Chen, Zhaoyang Zeng, Tianhe Ren, and Lei Zhang.
\newblock Chatrex: Taming multimodal llm for joint perception and understanding, 2024.

\bibitem[Koh et~al.(2023)Koh, Salakhutdinov, and Fried]{koh2023grounding}
Jing~Yu Koh, Ruslan Salakhutdinov, and Daniel Fried.
\newblock Grounding language models to images for multimodal inputs and outputs.
\newblock In \emph{International Conference on Machine Learning}, pages 17283--17300. PMLR, 2023.

\bibitem[Kool et~al.(2019)Kool, van Hoof, and Welling]{kool2019buy}
Wouter Kool, Herke van Hoof, and Max Welling.
\newblock Buy 4 reinforce samples, get a baseline for free!
\newblock 2019.

\bibitem[Lai et~al.(2024)Lai, Tian, Chen, Li, Yuan, Liu, and Jia]{lai2024lisa}
Xin Lai, Zhuotao Tian, Yukang Chen, Yanwei Li, Yuhui Yuan, Shu Liu, and Jiaya Jia.
\newblock Lisa: Reasoning segmentation via large language model.
\newblock In \emph{Proceedings of the IEEE/CVF Conference on Computer Vision and Pattern Recognition}, pages 9579--9589, 2024.

\bibitem[Li et~al.(2024)Li, Zhang, Guo, Zhang, Li, Zhang, Zhang, Zhang, Li, Liu, et~al.]{li2024llava}
Bo Li, Yuanhan Zhang, Dong Guo, Renrui Zhang, Feng Li, Hao Zhang, Kaichen Zhang, Peiyuan Zhang, Yanwei Li, Ziwei Liu, et~al.
\newblock Llava-onevision: Easy visual task transfer.
\newblock \emph{arXiv preprint arXiv:2408.03326}, 2024.

\bibitem[Li et~al.(2023)Li, Li, Savarese, and Hoi]{li2023blip}
Junnan Li, Dongxu Li, Silvio Savarese, and Steven Hoi.
\newblock Blip-2: Bootstrapping language-image pre-training with frozen image encoders and large language models.
\newblock In \emph{International conference on machine learning}, pages 19730--19742. PMLR, 2023.

\bibitem[Lin et~al.(2014)Lin, Maire, Belongie, Hays, Perona, Ramanan, Doll{\'a}r, and Zitnick]{lin2014microsoft}
Tsung-Yi Lin, Michael Maire, Serge Belongie, James Hays, Pietro Perona, Deva Ramanan, Piotr Doll{\'a}r, and C~Lawrence Zitnick.
\newblock Microsoft coco: Common objects in context.
\newblock In \emph{Computer vision--ECCV 2014: 13th European conference, zurich, Switzerland, September 6-12, 2014, proceedings, part v 13}, pages 740--755. Springer, 2014.

\bibitem[Liu et~al.(2024{\natexlab{a}})Liu, Zeng, Liu, Yan, He, Wang, Yan, Liu, and Zhou]{liu2024skywork}
Chris~Yuhao Liu, Liang Zeng, Jiacai Liu, Rui Yan, Jujie He, Chaojie Wang, Shuicheng Yan, Yang Liu, and Yahui Zhou.
\newblock Skywork-reward: Bag of tricks for reward modeling in llms.
\newblock \emph{arXiv preprint arXiv:2410.18451}, 2024{\natexlab{a}}.

\bibitem[Liu et~al.(2024{\natexlab{b}})Liu, Li, Li, and Lee]{liu2024improved}
Haotian Liu, Chunyuan Li, Yuheng Li, and Yong~Jae Lee.
\newblock Improved baselines with visual instruction tuning.
\newblock In \emph{Proceedings of the IEEE/CVF Conference on Computer Vision and Pattern Recognition}, pages 26296--26306, 2024{\natexlab{b}}.

\bibitem[Liu et~al.(2024{\natexlab{c}})Liu, Li, Li, Li, Zhang, Shen, and Lee]{liu2024llavanext}
Haotian Liu, Chunyuan Li, Yuheng Li, Bo Li, Yuanhan Zhang, Sheng Shen, and Yong~Jae Lee.
\newblock Llava-next: Improved reasoning, ocr, and world knowledge, 2024{\natexlab{c}}.

\bibitem[Liu et~al.(2024{\natexlab{d}})Liu, Li, Wu, and Lee]{liu2024visual}
Haotian Liu, Chunyuan Li, Qingyang Wu, and Yong~Jae Lee.
\newblock Visual instruction tuning.
\newblock \emph{Advances in neural information processing systems}, 36, 2024{\natexlab{d}}.

\bibitem[Liu et~al.(2024{\natexlab{e}})Liu, Zeng, Ren, Li, Zhang, Yang, Jiang, Li, Yang, Su, et~al.]{liu2024grounding}
Shilong Liu, Zhaoyang Zeng, Tianhe Ren, Feng Li, Hao Zhang, Jie Yang, Qing Jiang, Chunyuan Li, Jianwei Yang, Hang Su, et~al.
\newblock Grounding dino: Marrying dino with grounded pre-training for open-set object detection.
\newblock In \emph{European Conference on Computer Vision}, pages 38--55. Springer, 2024{\natexlab{e}}.

\bibitem[Liu et~al.(2023)Liu, Moosavi, and Lin]{liu2023llms}
Yiqi Liu, Nafise~Sadat Moosavi, and Chenghua Lin.
\newblock Llms as narcissistic evaluators: When ego inflates evaluation scores.
\newblock \emph{arXiv preprint arXiv:2311.09766}, 2023.

\bibitem[Liu et~al.(2025{\natexlab{a}})Liu, Chen, Li, Qi, Pang, Du, Lee, and Lin]{liu2025understanding}
Zichen Liu, Changyu Chen, Wenjun Li, Penghui Qi, Tianyu Pang, Chao Du, Wee~Sun Lee, and Min Lin.
\newblock Understanding r1-zero-like training: A critical perspective.
\newblock \emph{arXiv preprint arXiv:2503.20783}, 2025{\natexlab{a}}.

\bibitem[Liu et~al.(2025{\natexlab{b}})Liu, Sun, Zang, Dong, Cao, Duan, Lin, and Wang]{liu2025visual}
Ziyu Liu, Zeyi Sun, Yuhang Zang, Xiaoyi Dong, Yuhang Cao, Haodong Duan, Dahua Lin, and Jiaqi Wang.
\newblock Visual-rft: Visual reinforcement fine-tuning.
\newblock \emph{arXiv preprint arXiv:2503.01785}, 2025{\natexlab{b}}.

\bibitem[Lu et~al.(2023)Lu, Bansal, Xia, Liu, Li, Hajishirzi, Cheng, Chang, Galley, and Gao]{lu2023mathvista}
Pan Lu, Hritik Bansal, Tony Xia, Jiacheng Liu, Chunyuan Li, Hannaneh Hajishirzi, Hao Cheng, Kai-Wei Chang, Michel Galley, and Jianfeng Gao.
\newblock Mathvista: Evaluating mathematical reasoning of foundation models in visual contexts.
\newblock \emph{arXiv preprint arXiv:2310.02255}, 2023.

\bibitem[Mao et~al.(2016)Mao, Huang, Toshev, Camburu, Yuille, and Murphy]{mao2016generation}
Junhua Mao, Jonathan Huang, Alexander Toshev, Oana Camburu, Alan~L Yuille, and Kevin Murphy.
\newblock Generation and comprehension of unambiguous object descriptions.
\newblock In \emph{Proceedings of the IEEE conference on computer vision and pattern recognition}, pages 11--20, 2016.

\bibitem[Meng et~al.(2025)Meng, Du, Liu, Zhou, Lu, Fu, Shi, Wang, He, Zhang, et~al.]{meng2025mm}
Fanqing Meng, Lingxiao Du, Zongkai Liu, Zhixiang Zhou, Quanfeng Lu, Daocheng Fu, Botian Shi, Wenhai Wang, Junjun He, Kaipeng Zhang, et~al.
\newblock Mm-eureka: Exploring visual aha moment with rule-based large-scale reinforcement learning.
\newblock \emph{arXiv preprint arXiv:2503.07365}, 2025.

\bibitem[Ouyang et~al.(2022)Ouyang, Wu, Jiang, Almeida, Wainwright, Mishkin, Zhang, Agarwal, Slama, Ray, et~al.]{ouyang2022training}
Long Ouyang, Jeffrey Wu, Xu Jiang, Diogo Almeida, Carroll Wainwright, Pamela Mishkin, Chong Zhang, Sandhini Agarwal, Katarina Slama, Alex Ray, et~al.
\newblock Training language models to follow instructions with human feedback.
\newblock \emph{Advances in neural information processing systems}, 35:\penalty0 27730--27744, 2022.

\bibitem[Pan et~al.(2024{\natexlab{a}})Pan, Jones, Jagadeesan, and Steinhardt]{pan2024feedback}
Alexander Pan, Erik Jones, Meena Jagadeesan, and Jacob Steinhardt.
\newblock Feedback loops with language models drive in-context reward hacking.
\newblock \emph{arXiv preprint arXiv:2402.06627}, 2024{\natexlab{a}}.

\bibitem[Pan et~al.(2024{\natexlab{b}})Pan, He, Bowman, and Feng]{pan2024spontaneous}
Jane Pan, He He, Samuel~R Bowman, and Shi Feng.
\newblock Spontaneous reward hacking in iterative self-refinement.
\newblock \emph{arXiv preprint arXiv:2407.04549}, 2024{\natexlab{b}}.

\bibitem[Peng et~al.(2025)Peng, Zhang, Zhang, You, Liu, Zhu, Yang, Xu, Geng, and Yang]{peng2025lmm}
Yingzhe Peng, Gongrui Zhang, Miaosen Zhang, Zhiyuan You, Jie Liu, Qipeng Zhu, Kai Yang, Xingzhong Xu, Xin Geng, and Xu Yang.
\newblock Lmm-r1: Empowering 3b lmms with strong reasoning abilities through two-stage rule-based rl.
\newblock \emph{arXiv preprint arXiv:2503.07536}, 2025.

\bibitem[Radford et~al.(2021)Radford, Kim, Hallacy, Ramesh, Goh, Agarwal, Sastry, Askell, Mishkin, Clark, et~al.]{radford2021learning}
Alec Radford, Jong~Wook Kim, Chris Hallacy, Aditya Ramesh, Gabriel Goh, Sandhini Agarwal, Girish Sastry, Amanda Askell, Pamela Mishkin, Jack Clark, et~al.
\newblock Learning transferable visual models from natural language supervision.
\newblock In \emph{International conference on machine learning}, pages 8748--8763. PMLR, 2021.

\bibitem[Ren et~al.(2024)Ren, Jiang, Liu, Zeng, Liu, Gao, Huang, Ma, Jiang, Chen, et~al.]{ren2024grounding}
Tianhe Ren, Qing Jiang, Shilong Liu, Zhaoyang Zeng, Wenlong Liu, Han Gao, Hongjie Huang, Zhengyu Ma, Xiaoke Jiang, Yihao Chen, et~al.
\newblock Grounding dino 1.5: Advance the" edge" of open-set object detection.
\newblock \emph{arXiv preprint arXiv:2405.10300}, 2024.

\bibitem[Schulman et~al.(2017)Schulman, Wolski, Dhariwal, Radford, and Klimov]{schulman2017proximal}
John Schulman, Filip Wolski, Prafulla Dhariwal, Alec Radford, and Oleg Klimov.
\newblock Proximal policy optimization algorithms.
\newblock \emph{arXiv preprint arXiv:1707.06347}, 2017.

\bibitem[Shao et~al.(2024)Shao, Wang, Zhu, Xu, Song, Bi, Zhang, Zhang, Li, Wu, et~al.]{shao2024deepseekmath}
Zhihong Shao, Peiyi Wang, Qihao Zhu, Runxin Xu, Junxiao Song, Xiao Bi, Haowei Zhang, Mingchuan Zhang, YK Li, Y Wu, et~al.
\newblock Deepseekmath: Pushing the limits of mathematical reasoning in open language models.
\newblock \emph{arXiv preprint arXiv:2402.03300}, 2024.

\bibitem[Sun et~al.(2023)Sun, Fang, Wu, Wang, and Cao]{EVA-CLIP}
Quan Sun, Yuxin Fang, Ledell Wu, Xinlong Wang, and Yue Cao.
\newblock Eva-clip: Improved training techniques for clip at scale.
\newblock \emph{arXiv preprint arXiv:2303.15389}, 2023.

\bibitem[Wang et~al.(2024{\natexlab{a}})Wang, Pan, Shi, Lu, Ren, Zhou, Zhan, and Li]{wang2024measuring}
Ke Wang, Junting Pan, Weikang Shi, Zimu Lu, Houxing Ren, Aojun Zhou, Mingjie Zhan, and Hongsheng Li.
\newblock Measuring multimodal mathematical reasoning with math-vision dataset.
\newblock \emph{Advances in Neural Information Processing Systems}, 37:\penalty0 95095--95169, 2024{\natexlab{a}}.

\bibitem[Wang et~al.(2023)Wang, Li, Chen, Cai, Zhu, Lin, Cao, Liu, Liu, and Sui]{wang2023large}
Peiyi Wang, Lei Li, Liang Chen, Zefan Cai, Dawei Zhu, Binghuai Lin, Yunbo Cao, Qi Liu, Tianyu Liu, and Zhifang Sui.
\newblock Large language models are not fair evaluators.
\newblock \emph{arXiv preprint arXiv:2305.17926}, 2023.

\bibitem[Wang et~al.(2024{\natexlab{b}})Wang, Bai, Tan, Wang, Fan, Bai, Chen, Liu, Wang, Ge, et~al.]{wang2024qwen2}
Peng Wang, Shuai Bai, Sinan Tan, Shijie Wang, Zhihao Fan, Jinze Bai, Keqin Chen, Xuejing Liu, Jialin Wang, Wenbin Ge, et~al.
\newblock Qwen2-vl: Enhancing vision-language model's perception of the world at any resolution.
\newblock \emph{arXiv preprint arXiv:2409.12191}, 2024{\natexlab{b}}.

\bibitem[Wen et~al.(2024)Wen, Zhong, Khan, Perez, Steinhardt, Huang, Bowman, He, and Feng]{wen2024language}
Jiaxin Wen, Ruiqi Zhong, Akbir Khan, Ethan Perez, Jacob Steinhardt, Minlie Huang, Samuel~R Bowman, He He, and Shi Feng.
\newblock Language models learn to mislead humans via rlhf.
\newblock \emph{arXiv preprint arXiv:2409.12822}, 2024.

\bibitem[Xie et~al.(2023)Xie, Zhang, Wu, Zhu, Zhao, and Liang]{xie2023described}
Chi Xie, Zhao Zhang, Yixuan Wu, Feng Zhu, Rui Zhao, and Shuang Liang.
\newblock Described object detection: Liberating object detection with flexible expressions.
\newblock \emph{Advances in Neural Information Processing Systems}, 36:\penalty0 79095--79107, 2023.

\bibitem[Yang et~al.(2025)Yang, He, Pan, Jiang, Deng, Yang, Lu, Yin, Rao, Zhu, et~al.]{yang2025r1}
Yi Yang, Xiaoxuan He, Hongkun Pan, Xiyan Jiang, Yan Deng, Xingtao Yang, Haoyu Lu, Dacheng Yin, Fengyun Rao, Minfeng Zhu, et~al.
\newblock R1-onevision: Advancing generalized multimodal reasoning through cross-modal formalization.
\newblock \emph{arXiv preprint arXiv:2503.10615}, 2025.

\bibitem[Yao et~al.(2023)Yao, Liu, Zhao, Zhang, Liao, Fang, Lee, and Wang]{yao2023evaluate}
Yiyang Yao, Peng Liu, Tiancheng Zhao, Qianqian Zhang, Jiajia Liao, Chunxin Fang, Kyusong Lee, and Qing Wang.
\newblock How to evaluate the generalization of detection? a benchmark for comprehensive open-vocabulary detection.
\newblock \emph{arXiv preprint arXiv:2308.13177}, 2023.

\bibitem[Yu et~al.(2016)Yu, Poirson, Yang, Berg, and Berg]{yu2016modeling}
Licheng Yu, Patrick Poirson, Shan Yang, Alexander~C Berg, and Tamara~L Berg.
\newblock Modeling context in referring expressions.
\newblock In \emph{Computer Vision--ECCV 2016: 14th European Conference, Amsterdam, The Netherlands, October 11-14, 2016, Proceedings, Part II 14}, pages 69--85. Springer, 2016.

\bibitem[Zang et~al.(2025)Zang, Dong, Zhang, Cao, Liu, Ding, Wu, Ma, Duan, Zhang, et~al.]{zang2025internlm}
Yuhang Zang, Xiaoyi Dong, Pan Zhang, Yuhang Cao, Ziyu Liu, Shengyuan Ding, Shenxi Wu, Yubo Ma, Haodong Duan, Wenwei Zhang, et~al.
\newblock Internlm-xcomposer2. 5-reward: A simple yet effective multi-modal reward model.
\newblock \emph{arXiv preprint arXiv:2501.12368}, 2025.

\bibitem[Zhai et~al.(2023)Zhai, Mustafa, Kolesnikov, and Beyer]{zhai2023sigmoid}
Xiaohua Zhai, Basil Mustafa, Alexander Kolesnikov, and Lucas Beyer.
\newblock Sigmoid loss for language image pre-training, 2023.

\bibitem[Zhang et~al.(2024)Zhang, Jiang, Zhang, Lin, Guo, Qiu, Zhou, Lu, Chang, Qiao, et~al.]{zhang2024mathverse}
Renrui Zhang, Dongzhi Jiang, Yichi Zhang, Haokun Lin, Ziyu Guo, Pengshuo Qiu, Aojun Zhou, Pan Lu, Kai-Wei Chang, Yu Qiao, et~al.
\newblock Mathverse: Does your multi-modal llm truly see the diagrams in visual math problems?
\newblock In \emph{European Conference on Computer Vision}, pages 169--186. Springer, 2024.

\bibitem[Zhao et~al.(2022)Zhao, Liu, Lu, and Lee]{zhao2022omdet}
Tiancheng Zhao, Peng Liu, Xiaopeng Lu, and Kyusong Lee.
\newblock Omdet: Language-aware object detection with large-scale vision-language multi-dataset pre-training.
\newblock \emph{CoRR}, 2022.

\bibitem[Zhao et~al.(2024)Zhao, Liu, and Lee]{zhao2024omdet}
Tiancheng Zhao, Peng Liu, and Kyusong Lee.
\newblock Omdet: Large-scale vision-language multi-dataset pre-training with multimodal detection network.
\newblock \emph{IET Computer Vision}, 18\penalty0 (5):\penalty0 626--639, 2024.

\bibitem[Zhou et~al.(2025)Zhou, Li, Wang, Cheng, Zhou, and Hsieh]{zhou2025r1}
Hengguang Zhou, Xirui Li, Ruochen Wang, Minhao Cheng, Tianyi Zhou, and Cho-Jui Hsieh.
\newblock R1-zero's" aha moment" in visual reasoning on a 2b non-sft model.
\newblock \emph{arXiv preprint arXiv:2503.05132}, 2025.

\end{thebibliography}
